\documentclass{article} 

\usepackage{amsmath,amsfonts,bm}

\def\eqref#1{equation~\ref{#1}}

\def\1{\bm{1}}

\def\vb{{\bm{b}}}

\def\vd{{\bm{d}}}

\def\vr{{\bm{r}}}

\def\vu{{\bm{u}}}
\def\vv{{\bm{v}}}
\def\vw{{\bm{w}}}
\def\vx{{\bm{x}}}
\def\vy{{\bm{y}}}
\def\vz{{\bm{z}}}

\def\mE{{\bm{E}}}

\def\mW{{\bm{W}}}
\def\mX{{\bm{X}}}
\def\mY{{\bm{Y}}}

\DeclareMathAlphabet{\mathsfit}{\encodingdefault}{\sfdefault}{m}{sl}
\SetMathAlphabet{\mathsfit}{bold}{\encodingdefault}{\sfdefault}{bx}{n}

\def\sR{{\mathbb{R}}}

\usepackage{pgfplots}
\pgfplotsset{compat=1.18}
\usepackage{hyperref}
\usepackage{url}
\usepackage{pgfplots}
\usepackage{amsmath}
\usepackage{amsthm}
\usepackage{algorithm}
\usepackage{algpseudocode}
\usepackage{natbib}

\newcommand{\RR}{\ensuremath{\mathbb{R}}}
\newcommand{\NN}{\ensuremath{\mathbb{N}}}

\newtheorem{theorem}{Theorem}[section]
\newtheorem{proposition}[theorem]{Proposition}
\newtheorem{lemma}[theorem]{Lemma}
\newtheorem{corollary}[theorem]{Corollary}

\newtheorem{remark}[theorem]{Remark}

\begin{document}
  \title{A Parametric Linear Conic Projection for Structured Sparse Attention}
    \author{ Michel Barlaud$^1 $, Clement Genninasca$^1 $, Maxime Zanella$^2 $, \\ Benoit Macq$^2 $  and Beno\^it Legat$^2 $ \\
$^1 $ I3S Laboratory, CNRS, Universite Côte d'Azur, France \\
$^2 $ UCLouvain, Louvain-la-Neuve, Belgium.\\
}
\maketitle
\begin{abstract}
Structured sparsity has become a fundamental ingredient for scaling modern Transformer and Large Language Model (LLM) architectures, where dense attention matrices remain one of the primary bottlenecks in both memory footprint and computational cost. While many existing approaches rely on heuristic pruning or fixed sparse attention patterns, they generally lack a principled geometric formulation and strong optimization guarantees.

In this paper,
we introduce the Cone Alignment Index (CAI), a scale-invariant sparsity measure associated with a Lorentz-cone geometry.
We provide a probabilistic analysis of the distribution of the CAI score, including a general theorem and explicit corollaries for Gaussian and uniform distributions. As a consequence, we show that, in the high-sparsity regime, the probability that the samples lie within the convex cone tends to one.
Building upon this formulation, we derive two complementary projections: the \emph{Parametric Linear Conic Projection} (PLCP), an efficient closed-form approximation, and the \emph{Parametric Euclidean Conic Projection} (PECP), the exact Euclidean projection onto the cone boundary.

Our analysis establishes several fundamental theoretical properties. We prove that PLCP and PECP are collinear, differ only by a positive radial scaling factor, and induce exactly the same sparsity threshold, thereby preserving an identical active support. Furthermore, we derive a support-identification theorem showing that, for any prescribed sparsity level $l<n$, the number of coefficients discarded at each iteration is theoretically controlled.

These results lead to a simple and scalable two-stage algorithm. First, the active support is identified through a provably correct thresholding rule. Second, the final projection is obtained by a closed-form support-aware extrapolation coefficient, yielding an efficient bilevel projection algorithm whose computational complexity is essentially linear in the active support size.

Building on this theoretical foundation, we develop a bilevel PLCP framework for structured matrix sparsification. The proposed method naturally generates hardware-friendly structured sparsity patterns, including row-wise, column-wise, and diagonal masks, making it particularly well suited for accelerating Transformer attention mechanisms.

Experiments on large synthetic matrices and Transformer attention layers validate both the theoretical and practical advantages of the proposed approach. On GPUs, PLCP achieves a computational complexity comparable to structured pruning while providing up to a quadratic speed-up over GSP-Hybrid. On GLUE and SuperGLUE benchmarks, PLCP reaches up to $90\%$ attention sparsity with only minor accuracy degradation, consistently outperforming GSP, structured pruning, and fixed sparse-attention architectures such as BigBird at high sparsity levels.

Beyond sparse attention, the proposed framework provides a general geometric and optimization-based methodology for structured matrix sparsification, opening promising directions toward scalable, hardware-aware, and energy-efficient deep learning.

\end{abstract}

\section{State of the art of neural network sparsification}

\subsection{Challenge and classical methods.}
Many modern machine learning systems rely on the manipulation of large matrices whose computational and memory costs dominate both training and inference. Examples include weight matrices in deep neural networks, covariance matrices in inverse problems, and, more recently, attention matrices in transformer architectures Vaswani et al. (2017). As these matrices continue to grow in size, structured sparsity has emerged as a key mechanism for reducing computational complexity, memory footprint, and energy consumption \cite{Carbon,AIred}.
A central challenge is that effective sparsity mechanisms should simultaneously satisfy several desirable properties. First, they should provide explicit control over the sparsity level. Second, they should remain invariant to scale, since many optimization problems are unaffected by a global rescaling of the variables. Third, they should induce structured sparsity patterns that can be efficiently exploited by modern hardware architectures. Finally, the associated projection or optimization procedures should remain computationally efficient even for very large matrices.\\
To address these challenges, a large body of research has focused on neural network
sparsification, i.e., reducing the number of nonzero weights in a model.
We distinguish three main classes of methods for sparsifying attention matrices: pruning methods, architecture-based methods, and projection-based methods.
One of the earliest and most widely adopted approaches to induce sparsity in neural
networks is pruning \cite{Han2015,Alv2016,double,chen-Nips}.
Classical pruning methods \cite{Sanh}, RigL \cite{evci}, and SparseGPT \cite{Sparse-GPT}
eliminate weights using magnitude-based or gradient-based heuristics \cite{Sparsity}.
While effective in practice, these approaches yield empirical sparsity without enforcing
explicit geometric constraints.
More recent structured pruning methods \cite{LLaMA,Delete} attempt to overcome some of
the inefficiencies of classical pruning techniques.\\
Among large-scale matrix applications, attention matrices in transformer architectures constitute a particularly important use case. Sparse transformer architectures such as BigBird \cite{Bigbird}  reduce attention complexity through predefined sparse patterns including local windows, block structures, and random-global connections. While computationally efficient, these masks remain largely handcrafted and may fail to adapt to the structure of the data. Note that BigBird was designed for large attention matrices.

The Least Absolute Shrinkage and Selection Operator (LASSO) \cite{tRS,Sparse} promotes
sparsity through an $\ell_1$-norm penalty.
The $\ell_0$ norm, which directly counts the number of nonzero weights, provides exact
sparsity control and is scale-invariant, but it is non-differentiable \cite{L0}. 
$\ell_1$ projection-based approaches directly enforce sparsity by projecting weight
vectors onto norm balls, typically the $\ell_1$ ball, using efficient algorithms
\cite{duchi,condat}.
While such methods enjoy linear-time complexity, they are not scale-invariant and do not
naturally induce structured sparsity.
Since many modern hardware architectures implement multiply--add operations as a single
instruction, such irregular sparsity patterns fail to translate into practical speed-ups and are poorly suited for efficient parallel execution.\\

\subsection{Background on structured sparsity.}
A key limitation of pruning-based methods, as well as $\ell_1$- and $\ell_0$-induced
sparsity, is their unstructured nature, which typically produces randomly distributed
zero-valued weights. To overcome the inefficiency of unstructured sparsity, research has increasingly turned
toward structured sparsity methods \cite{structured}, which aim to remove entire groups of parameters, such
as filters, channels, or neurons.
Group LASSO and its variants introduce regularizers that enforce sparsity at the group
level \cite{Yua,group_lasso,Group-sparse,Group-ex,Sparse-groupLasso,Ma2019,GSP,chen2021}.
Of particular interest is the $\ell_{1,\infty}$ projection, which promotes structured
sparsity by enforcing group-wise shrinkage, for instance, by setting entire columns of a
weight matrix to zero.
Recent works have proposed efficient $\ell_{1,\infty}$ projection algorithms based on proximal methods \cite{Prox}.
However, their worst-case time complexity remains
$\mathcal{O}(nm \log(nm))$ for a matrix in $\mathbb{R}^{n\times m}$, which may limit their
scalability to very large neural networks.\\
An alternative is the Hoyer score \cite{hoyer}, which balances sparsity and
scale invariance.
It has been successfully applied in a variety of contexts, including blind
deconvolution \cite{Repetti_2015}, nonnegative least squares
\cite{esser2013,Gillis}, neural network regularization
\cite{Deep-Hoyer,GSP,Thom}. Nevertheless, these existing projections remain non-convex and typically require iterative numerical procedures.\\

Motivated by these limitations, we investigate whether a scale-invariant geometric constraint can provide a unified framework for structured matrix sparsification. Our objective is to derive a projection operator that simultaneously admits a closed-form solution, promotes structured sparsity, and remains computationally efficient at large scale. As a representative application, we demonstrate its effectiveness for sparsifying transformer attention matrices, while the proposed framework naturally extends to broader classes of matrix sparsification problems, including neural network parameter matrices and inverse problems.

\subsection{Contributions and Organization of this Work}
In summary, this paper makes the following contributions:

\begin{itemize}

\item We introduce the \emph{Cone Alignment Index} (CAI), a novel sparsity constraint whose level sets naturally define a Lorentz-cone geometry. Based on this formulation, we develop a new conic projection framework through a parametric optimization model.
\item We provide a probabilistic analysis of the distribution of the CAI score, including a general theorem and explicit corollaries for Gaussian and uniform distributions. As a consequence, we show that, in the high-sparsity regime, the probability that the samples lie within the convex cone tends to one.

\item We derive two complementary projections: the \emph{Parametric Linear Conic Projection} (PLCP), an efficient closed-form approximation, and the \emph{Parametric Euclidean Conic Projection} (PECP), the exact Euclidean projection onto the cone boundary. We prove that PLCP and PECP are collinear, differ only by a positive radial scaling factor, and induce exactly the same sparsity threshold and active support. 

\item Building upon these theoretical results, we propose a closed-form conic projection algorithm requiring only a single support-aware extrapolation step after support identification. We further extend this framework to structured matrix sparsification through a bilevel projection strategy that naturally generates hardware-friendly row-wise, column-wise, and diagonal sparsity patterns.

\item We perform an experimental evaluation on large-scale matrix sparsification and Transformer attention layers. PLCP is benchmarked against Group Lasso projections, structured pruning, GSP, and the widely used sparse-attention baseline BigBird. The proposed method achieves computational complexity comparable to structured pruning while providing up to a quadratic speed-up over GSP-Hybrid, and consistently delivers a superior accuracy-sparsity trade-off on GLUE and SuperGLUE benchmarks.

\end{itemize}

\section{A new Cone Alignment Index (CAI)}
We use the mathematical notation file \emph{math-commands.tex} from
the textbook, Deep Learning Goodfellow et al. (2016).

\subsection{Definition and Geometric Interpretation }

We define the \emph{Cone Alignment Index (CAI)} constraint for a vector $\vx \in \mathbb{R}^n_+$ (nonnegative orthant) as
\begin{equation}
\label{CAI}
CAI(\vx)
= \frac{(\sum_{i=1}^n x_i)^2}{\sum_{i=1}^n x_i^2}
= \frac{\|\vx\|_1^2}{\|\vx\|_2^2}
= \frac{(\mathbf{1}^\top \vx)^2}{\vx^\top \vx} \leq l
\end{equation}
where $\mathbf{1}$ denotes the all-ones vector in $\mathbb{R}^n$ and $l $ is the constraint level.
Note that, in the nonnegative orthant, the ratio
\[
CAI(x)=\frac{(\mathbf{1}^\top x)^2}{\|x\|_2^2}
\]
differs from the Hoyer score \cite{hoyer}, while being closely related to the scale-invariant sparsity regularization term introduced in \cite{Deep-Hoyer}. By construction, $CAI$ is \textbf{scale-invariant}.

\begin{lemma}[Geometric structure]
Consider now the constraint in Equation \ref{Cone}. It
defines a homogeneous convex cone when restricted to the nonnegative orthant.
\begin{equation}
\label{Cone}
\mathcal{C}_{l,in}^+
:=
\left\{
x\in\mathbb{R}_+^n
:\;
(\mathbf{1}^\top x)^2 > l\|x\|_2^2
\right\},
\quad l\in[1,n].
\end{equation}

This equation corresponds to the interior of a revolution cone with apex at the origin and axis along the diagonal direction $\mathbf{1}=(1,1,\dots,1)$  and aperture angle 
$\delta = \arccos(\sqrt{l/n})$.\\   
The interior of the cone is convex, whereas its boundary and exterior (Equation \ref{Bound}) are not:
\begin{equation}
\label{Bound}
\mathcal{C}_{l,out}^+
:=
\left\{
x\in\mathbb{R}_+^n
:\;
(\mathbf{1}^\top x)^2 \le l\|x\|_2^2
\right\},
\quad l\in[1,n].
\end{equation}
Note that $\mathcal{C}_{l,out}^+$ is precisely the set satisfying the CAI constraint of Equation \ref{CAI}. The two regions are complementary within $\mathcal{R}_{+}^{n}$: $\mathcal{C}_{l,in}^+$ is the (convex) region violating the constraint, while $\mathcal{C}_{l,out}^+$ is the (non-convex) feasible region, i.e., $\mathcal{C}_{l,out}^+ = \mathcal{R}_{+}^{n} \setminus \mathcal{C}_{l,in}^+$ 
\end{lemma}
For $l\in[1,n]$, the quantity $\mathrm{CAI}(\mathbf{x})$ measures
the alignment of $\mathbf{x}$ with the diagonal axis:
$\mathrm{CAI}(\mathbf{x})=n$ when $\mathbf{x}$ is collinear with
$\mathbf{1}$, whereas its minimum value $\mathrm{CAI}(\mathbf{x})=1$
is attained for a one-sparse vector.

\subsection{Geometric Interpretation of the CAI Projection}
\label{hypercone}
To compute the CAI projection, we first map the original data
$\vy^0 \in \mathbb{R}^n$ to the nonnegative orthant $\vy \in \mathbb{R}^n_+$ by setting
\[
y_i \leftarrow |y_i^0|, \qquad \forall i \in \{1,\dots,n\}.
\]
We then distinguish two cases depending on the position of $\vy$
with respect to the CAI cone.\\

\textbf{Case 1: Points satisfying the constraint.}
If $\vy$ satisfies the constraint  $(\mathbf{1}^\top x)^2 \le l\|x\|_2^2$ then $\vy$
already lies in the feasible set and no projection is required.
In this case, the point remains unchanged.\\
\textbf{Case 2: Points violating the constraint.}
If $(\mathbf{1}^\top x)^2 > l\|x\|_2^2$ then $\vy$ lies strictly inside the CAI cone,
which is a convex set, and must be projected onto its boundary.

\subsection{Distribution of the CAI score }
The objective of this section is to investigate the properties of the CAI score distribution.

\begin{theorem}[Asymptotic distribution of the CAI score]
\label{hoyer_general}
Let $Y_1,\ldots,Y_n$ be i.i.d.\ random variables such that
\[
0<\mu_2=\mathbb{E}[Y^2]<\infty,
\qquad
\mathbb{E}[Y^4]<\infty,
\]
and define
\[
H_n
=
\frac{\left(\sum_{i=1}^n |Y_i|\right)^2}
{\sum_{i=1}^n Y_i^2}.
\]

Let
\[
\mu_1=\mathbb{E}[|Y|],
\qquad
\mu_3=\mathbb{E}[|Y|^3],
\qquad
\mu_4=\mathbb{E}[Y^4],
\qquad
m=\frac{\mu_1^2}{\mu_2}.
\]

Then
\[
\frac{H_n}{n}\xrightarrow{P}m.
\]

Moreover,
\[
\sqrt{n}
\left(
\frac{H_n}{n}-m
\right)
\xrightarrow{d}
\mathcal{N}(0,V_H),
\]
where
\[
V_H
=
\frac{4\mu_1^2}{\mu_2^2}
\left(\mu_2-\mu_1^2\right)
+
\frac{\mu_1^4}{\mu_2^4}
\left(\mu_4-\mu_2^2\right)
-
\frac{4\mu_1^3}{\mu_2^3}
\left(\mu_3-\mu_1\mu_2\right).
\]

Therefore, for large $n$,
\begin{equation}
\label{Theo-general}
\boxed{
\frac{H_n}{n}
\approx
\mathcal{N}
\left(
m,\frac{V_H}{n}
\right).
}
\end{equation}

\end{theorem}

See proof in Appendix \ref{distribution}

\begin{corollary}[Gaussian case]
\label{hoyer_gaussian}
Let $\mathbf Y=(Y_1,\ldots,Y_n)$ with
$Y_i\stackrel{\mathrm{i.i.d.}}{\sim}\mathcal N(0,1)$.
Then
\[
H(\mathbf Y)
=
\frac{\|\mathbf Y\|_1^2}{\|\mathbf Y\|_2^2}
=
\|\mathbf Z\|_1^2,
\]
where
\[
\mathbf Z=\frac{\mathbf Y}{\|\mathbf Y\|_2}
\]
is uniformly distributed on the unit sphere.

Then
\[
\mu_1=\sqrt{\frac{2}{\pi}},
\qquad
\mu_2=1
\]
Hence
\[
m=\frac{\mu_1^2}{\mu_2}
=\frac{2}{\pi},
\]
and
\[
\sigma_H^2
=
\frac{8}{\pi}
\left(1-\frac{3}{\pi}\right).
\]

The linear-in-$n$ parametrization naturally follows from the relation between
the CAI parameter $l$ and the normalized sparsity. Indeed, for a vector of
dimension $n$, the CAI parameter satisfies $1\leq l\leq n$, and the normalized
sparsity level is approximately characterized by the dimensionless ratio
\[
s \simeq 1- \frac{l}{n}.
\]
Keeping a prescribed normalized sparsity level as the dimension
increases requires $l$ to scale linearly with $n$. This motivates the use of
$l/n$ as the natural dimension-independent parameter for comparing sparsity
regimes across different dimensions.
Therefore the normalized distribution is given by:
\[
\boxed{
\frac{H_n}{n}
\approx
\mathcal N
\left(
\frac{2}{\pi},
\frac{8}{\pi n}
\left(1-\frac{3}{\pi}\right)
\right).
}
\]

\textbf{Gaussian high-sparsity regime}
Applying Theorem \ref{hoyer_general}, to i.i.d.\ Gaussian samples, we obtain the following high-sparsity regime:
\[
\frac{l}{n}<\frac{2}{\pi}
\]
with a fixed positive margin, then
\[
\mathbb P(H_n>l)\longrightarrow 1
\qquad\text{as }n\to\infty.
\]
Consequently, almost all samples lie inside the convex CAI cone with probability tending to one in the high-sparsity regime
\[
\frac{l}{n}\ll\frac{2}{\pi},
\]
\end{corollary}

 \begin{figure}[ht]
   \centering
  \includegraphics[width=0.7\textwidth]{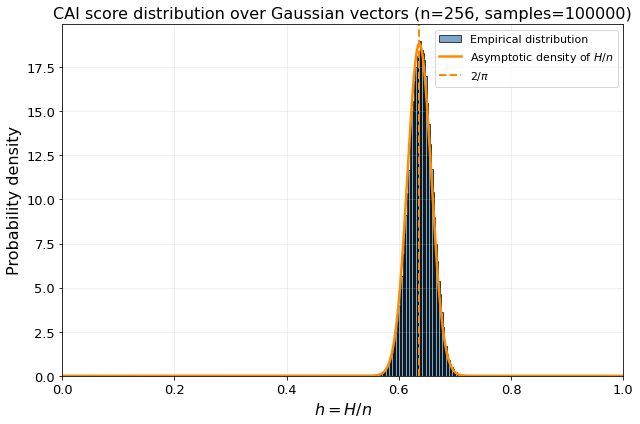}
   \caption{CAI score asymptotic distribution.}
  \label{density} 
\end{figure}

Figure \ref{density} shows the excellent agreement between the asymptotic distribution and the empirical distribution obtained from Gaussian data.

\begin{corollary}[Uniform distribution]
\label{hoyer_uniform}

Let $Y_1,\ldots,Y_n$ be i.i.d. random variables with
\[
Y_i \sim \mathcal U(0,1).
\]
Then
\[
m=\frac{\mu_1^2}{\mu_2}=\frac34.
\]
Moreover, for sufficiently large $n$,
\[
\frac{H_n}{n}
\approx
\mathcal N\left(\frac34,\frac{3}{40n}\right).
\]
Equivalently,
\[
H_n
\approx
\mathcal N\left(\frac{3n}{4},\frac{3n}{40}\right).
\]

\textbf{Uniform high-sparsity regime}
For i.i.d.\ uniformly distributed samples, if
\[
\frac{l}{n}<\frac34
\]
with a fixed positive margin, then
\[
\mathbb P(H_n>l)\longrightarrow 1
\qquad\text{as }n\to\infty.
\]
Consequently, asymptotically, the samples lie inside the convex CAI cone
and violate the target sparsity constraint $H_n\leq l$.

In particular, in the high-sparsity regime
\[
\frac{l}{n}\ll\frac34,
\]
almost all samples lie inside the CAI cone with probability tending to one.
\end{corollary}

\begin{remark}
Note that the random variable $H(Y)$ where $Y \sim N(0, 1)$ can be equivalently defined as the L1 norm $\|Z\|_1^2$ of a vector $Z$ sampled uniformly over the unit sphere.
This relates our result to \cite[Theorem~3.6]{kabluchko2016meanwidthregularpolytopes} and the particular case for $n = 1$ of \cite[Theorem~1.1]{paouris2012centrallimittheoremprojections} which studies $\|Z\|_1$.\\
Since Theorem~2.2 considers the square of $\|Z\|_1$, our result
can be derived from theirs by applying the delta method.\\
For large $n$, the second-order correction to the mean of the squared
CAI score becomes negligible compared with the leading-order term.
Moreover, the second-order shift also becomes negligible compared
with the standard deviation when $n$ is large. 
\end{remark}
See proof in Appendix \ref{second}

\section{ PLCP and PECP with Common Active-Set Selection}


\subsection{Sparse Active-Set Selection}
\label{Sec-threshold}

We introduce a unified formulation of a Parametric Linear Conic Projection (PLCP) and a Parametric Euclidean Conic Projection  (PECP).
Both methods share the same CAI geometry and the same active-set
selection procedure. In particular, they use the same threshold
$\alpha$ and therefore identify the same active support.

They differ only in the final geometric projection. PLCP uses an affine
parametric construction characterized by the extrapolation factor
$\lambda$, whereas PECP performs a Euclidean projection characterized
by the parameter $t$, or equivalently by the radial scaling factor
$\kappa$.
Let denote the CAI score of the original input by
\[
\boxed{
H_0=H(\vy).
}
\]

Since
\[
H(\vy)=H(|\vy|),
\]
the active-set selection can be performed on the nonnegative vector
\[
\vx^{(0)}=|\vy|.
\]

The overall construction is summarized by
\[
\boxed{
\text{CAI geometry}
\longrightarrow
\text{common active-set selection}
\longrightarrow
\begin{cases}
\text{PLCP projection},\\
\text{PECP projection}.
\end{cases}
}
\]

The active-set threshold is obtained directly from the CAI boundary
condition

At iteration $k$, let
\[
\mathcal A_k
=
\left\{
i:x_i^{(k)}>0
\right\}
\]
denote the current active support, with
\[
\nu_k
=
|\mathcal A_k|
=
\|\vx^{(k)}\|_0.
\]

Consider the shifted vector on the current active support
\[
\vz^{(k)}
=
\vx^{(k)}
-
\alpha_k\mathbf 1_{\mathcal A_k}.
\]

The threshold $\alpha_k$ is determined by requiring the shifted vector
to lie on the CAI boundary:
\[
\boxed{
H(\vz^{(k)})=l.
}
\]

The threshold is given by: (See Appendix \ref{active-set})
\begin{equation}
\alpha_k
=
\frac{\|\vx^{(k)}\|_1}{\nu_k}
\left(
1-
\sqrt{
\frac{
l\left(\nu_k-H(\vx^{(k)})\right)
}{
H(\vx^{(k)})(\nu_k-l)
}
}
\right).
\end{equation}

The threshold $\alpha_k$ is therefore a geometric threshold induced by
the CAI boundary.

The active support is updated according to
\[
x_i^{(k+1)}
=
\begin{cases}
x_i^{(k)}, & x_i^{(k)}\geq\alpha_k,\\
0,         & x_i^{(k)}<\alpha_k.
\end{cases}
\]

Thus the vector becomes progressively sparser across iterations (see Sparsity analysis in section \ref{sparsity}).
The active-set iterations are repeated until the support stabilizes (See convergence analysis in section \ref{convergence})

\begin{lemma}
Let $\vx$ be the final active-set vector, the last
threshold $\alpha$ computed before stabilization is :

\begin{equation}
\label{Threshold}
\boxed{
\alpha
=
\frac{\|\vx\|_1}{\nu}
\left(
1-
\sqrt{
\frac{
l\left(\nu-H(\vx)\right)
}{
H(\vx)(\nu-l)
}
}
\right).
}   
\end{equation}
\end{lemma}

\begin{proposition}[Lower bound on the active support]
\label{support_bound}
Equation~(6) becomes singular when $\nu=l$, since its denominator
vanishes. However, the CAI boundary condition imposes a natural lower
bound on the admissible support size.

 Let $\vx\in\RR_+^n$ with support $S$ and $|S|=\nu$.
By Cauchy-Schwarz applied to $S$,
\[
(\mathbf{1}^\top \vx)^2=\Big(\sum_{i\in S}x_i\Big)^2
\le \Big(\sum_{i\in S}1^2\Big)\Big(\sum_{i\in S}x_i^2\Big)
= \nu\,\|\vx\|_2^2.
\]
Hence, dividing both sides by $\sum_{i\in S}x_i^2=\|x\|_2^2$ yields $CAI(x)\le \nu$.\\
If $x\in\RR_+^n$ satisfies the cone-surface constraint, $CAI(\vx)= l$,
then the support size $\nu$ necessarily satisfies $l\le \nu$.\\
Moreover, the support size $ \nu \in \NN$, thus in particular, if 
\begin{equation}
\boxed{
l\notin\NN,\quad then \quad \nu > l.}
\end{equation}
This result establishes a fundamental lower bound on the active support size and guarantees that the support-selection procedure using Equation \ref{Threshold} in Algorithm 1 cannot collapse below this bound.
\end{proposition}

\subsection{PLCP projection}
\label{sec:lambda}
For an original input vector satisfying
\[
H_0=H(\vy)>l,
\] PLCP constructs the CAI boundary solution
through the following affine parametrization
\[
\boxed{
\widehat{\vx}_{\mathrm{PLCP}}
=
\lambda\vx
+
(1-\lambda)\vd.
}
\]
with 
\[
\mathbf d
=
\frac{\|\mathbf x\|_1}{\nu}\mathbf 1.
\]

\begin{lemma}

Let $\vx$ be the final active-set vector,
the PLCP extrapolation factor $ \lambda$
is directly obtained from the final threshold $\alpha$.
(see proof in section \ref{PLCP_lambda})

\[
\boxed{
\lambda
=
\frac{1}
{1-\dfrac{\alpha\nu}{\|\vx\|_1}}.
}
\]

Equivalently,
\[
\boxed{
\lambda
=
\sqrt{
\frac{
H(\vx)(\nu-l)
}{
l\left(\nu-H(\vx)\right)
}
}.
}
\]
\end{lemma}
Thus, $\alpha$ and $\lambda$ are not independent parameters. Once the
active support has stabilized and $\alpha$ is known, $\lambda$ follows
immediately.

Thus, the stabilized active-set quantities determine both the
extrapolation factor and the scale of the diagonal reference vector.
The PLCP construction can therefore be summarized as
\[
\boxed{
H,l
\longrightarrow
\alpha
\longrightarrow
\vx 
\longrightarrow
\lambda
\longrightarrow
\widehat{\vx}_{\mathrm{PLCP}}
}
\]

PLCP is therefore a closed-form parametric construction satisfying the
CAI geometry.

\subsection{PECP Euclidean-norm Projection}
\label{pecp_projection}

PECP uses the same active support $\mathcal A$ identified by the common
thresholding procedure. PECP selects the point on this CAI boundary generator that minimizes the Euclidean distance to the input vector.

For an original input vector satisfying $H_0=H(\vy)>l$,
PECP seeks, on the active support $A$ identified by the common
thresholding procedure, the closest point on the corresponding
CAI boundary:
\[
\widehat{\vx}_{\rm PECP}
=
\arg\min_{\vx}
\frac12\|\vx-\vy_A\|_2^2
\qquad
\text{subject to}
\qquad H(\vx)=l.
\]

On the final active support of size $\nu$, define
\[
\mathbf u_\nu=\frac{1}{\sqrt{\nu}}\mathbf 1_A.
\]
The input vector restricted to the active support is decomposed as
\[
\mathbf y_A=s\mathbf u_\nu+r\mathbf v,
\]
where
\[
s=\mathbf y_A^\top\mathbf u_\nu
=\frac{\|\mathbf y_A\|_1}{\sqrt{\nu}},
\qquad
r=\|\mathbf y_A-s\mathbf u_\nu\|_2.
\]
For $r>0$, define
\[
\mathbf v=
\frac{\mathbf y_A-s\mathbf u_\nu}{r}.
\]

Hence, using these notations, PECP reduces to the one-dimensional minimization problem
(see details in Appendix \ref{PECPfrob})

\[
\boxed{
\arg\min_{t\geq0}
\;
(s-t)^2+(r-tq)^2.
}
\]

\begin{lemma}
Differentiating with respect to $t$,
the Euclidean projection onto the CAI boundary is 
\begin{equation}
\boxed{
\widehat{\vx}_{\mathrm{PECP}}
=
t\vu+t q\vv,
}
\end{equation}
where
\[
\boxed{
t
=
\frac{s+qr}{1+q^2}.
}
\]

The PECP construction can therefore be summarized as
\[
\boxed{
H,l
\longrightarrow
\alpha
\longrightarrow
\vx
\longrightarrow
t
\longrightarrow
\widehat{\vx}_{\mathrm{PECP}}
}
\]

Hence, the PECP solution is the Euclidean projection onto the
corresponding CAI boundary generator. In the $(s,r)$ plane, this
reduces to the orthogonal projection of the point $(s,r)$ onto the
boundary line $r=qs$.

\end{lemma}

\paragraph{Collinearity between PLCP and PECP}

PLCP and PECP use the same active support and reach the same generator
of the CAI boundary.
Consequently, their boundary solutions are collinear: (See proof in section \ref{Euclide})
\[
\boxed{
\widehat{\vx}_{\mathrm{PECP}}
=
\kappa
\widehat{\vx}_{\mathrm{PLCP}}.
}
\]

with
\begin{equation}
\boxed{
\kappa
=
\frac{t}{s_{\mathrm{PLCP}}}.
}
\end{equation}


\begin{algorithm}[H]
\caption{ Iterative PLCP}
\label{PLCP}
\begin{algorithmic}[1]
\Require $\vy\in\mathbb R^n$, target parameter $l$
\Ensure $\widehat{\vx}\in\mathbb R^n$
\State $H_0\gets H(\vy)$
\State $x_i\gets|y_i|$,
\quad $\forall i\in\{1,\ldots,n\}$

\State $\nu\gets\ell_0(\vx)+1$

\While{$\nu\neq\ell_0(\vx)$}

    \State $\nu\gets\ell_0(\vx)$

    \State
    $\displaystyle
    \alpha
    \gets
    \frac{\|\vx\|_1}{\nu}
    \left(
    1-
    \sqrt{
    \frac{ 
    l\left(\nu-H(\vx)\right)
    }{
    H(\vx)(\nu-l)
    }}
    \right)
    $

    \State
    $\displaystyle
    x_i
    \gets
    \begin{cases}
    x_i, & x_i\geq\alpha,\\
    0,   & x_i<\alpha,
    \end{cases}
    \qquad
    \forall i\in\{1,\ldots,n\}$

\EndWhile

\If{$H_0>l$}

    \State
    $\displaystyle
    \lambda
    \gets
    \frac{1}
    {1-\dfrac{\alpha\nu}{\|\vx\|_1}}
    $

    \State
    $\displaystyle
    d_i \leftarrow
    \begin{cases}
    \dfrac{\|\mathbf x\|_1}{\nu}, & x_i\neq 0,\\[1mm]
    0, & x_i=0.
    \end{cases}
    \qquad
    \forall i$

    \State
    $\widehat{\vx}
    \gets
    \lambda\vx+(1-\lambda)\vd$

\Else

    \State $\widehat{\vx}\gets\vx$

\EndIf

\State
$\widehat{x}_i
\gets
\widehat{x}_i\operatorname{sign}(y_i),\quad \forall i$
\State \Return $\widehat{\vx}$
\end{algorithmic}
\end{algorithm}

\begin{remark}
Since Algorithms~\ref{PLCP} and~\ref{PECP} use the same initialization and the same thresholding rule, they generate identical active-set iterates and therefore identify the same final active support. On this support, their respective projections lie on the same CAI boundary generator; hence, their final outputs are positively collinear. Therefore, PLCP preserves the direction and the support pattern of the Euclidean projection, while generally differing in radial scaling. We consequently refer to PLCP as a parametric conic projection.
\end{remark}


\begin{algorithm}[H]
\caption{Iterative PECP}
\label{PECP}
\begin{algorithmic}[1]
\Require $\vy\in\mathbb R^n$, target parameter $l$
\Ensure $\widehat{\vx}\in\mathbb R^n$
\State $H_0\gets H(\vy)$
\State $x_i\gets|y_i|$,
\quad $\forall i\in\{1,\ldots,n\}$
\State $\nu\gets\ell_0(\vx)+1$
\While{$\nu\neq\ell_0(\vx)$}
    \State $\nu\gets\ell_0(\vx)$
    \State
    $\displaystyle
    \alpha
    \gets
    \frac{\|\vx\|_1}{\nu}
    \left(
    1-
    \sqrt{
    \frac{
    l\left(\nu-H(\vx)\right)
    }{
    H(\vx)(\nu-l)
    }}
    \right)
    $

    \State
    $\displaystyle
    x_i
    \gets
    \begin{cases}
    x_i, & x_i\geq\alpha,\\
    0,   & x_i<\alpha,
    \end{cases}
    \qquad
    \forall i$

\EndWhile

\If{$H_0>l$}
    \State $\mathcal A\gets\{i:x_i\neq0\}$
    \State $\nu\gets|\mathcal A|$
    \State
    $\displaystyle
    \vu_\nu
    \gets
    \frac{1}{\sqrt{\nu}}
    \mathbf 1_{\mathcal A}
    $

    \State
    $\displaystyle
    s
    \gets
    \vx^\top\vu_\nu
    =
    \frac{\|\vx\|_1}{\sqrt{\nu}}
    $

    \State
    $\displaystyle
    r
    \gets
    \|\vx-s\vu_\nu\|_2
    $

    \State
    $\displaystyle
    q
    \gets
    \sqrt{\frac{\nu-l}{l}}
    $
    \State
    $\displaystyle
    t
    \gets
    \frac{s+qr}{1+q^2}
    $

    \If{$r>0$}

        \State
        $\displaystyle
        \vv
        \gets
        \frac{\vx-s\vu_\nu}{r}
        $

        \State
        $\displaystyle
        \widehat{\vx}
        \gets
        t\vu_\nu+tq\vv
        $

    \EndIf

\Else

    \State $\widehat{\vx}\gets\vx$

\EndIf

\State
$\widehat{x}_i
\gets
\widehat{x}_i\operatorname{sign}(y_i),
\quad \forall i$
\State \Return $\widehat{\vx}$
\end{algorithmic}
\end{algorithm}

\begin{remark} 
Note that the test $H_0>l$ in line 9 for PLCP and line 9 for PECP is unnecessary in the case of the high-sparsity regime.\\

\end{remark}

\subsection{Sparsity analysis of the active set selection}
\label{sparsity}
\begin{theorem}
\label{Jensen}
Let $y_1,\dots,y_n$ be positive real numbers, and set
\[
 l_1:=\sum_{i=1}^n y_i,
 \qquad
 l_2:=\sqrt{\sum_{i=1}^n y_i^2},
 \qquad
 H:=\left(\frac{l_1}{l_2}\right)^2.
\]
Define
\[
\mu:=\frac{l_1}{n},
\qquad
\sigma^2:=\frac1n\sum_{i=1}^n (y_i-\mu)^2.
\]
Fix a positive number $l<n$. Then the number of indices $i$ such that
\begin{equation}
\boxed{
 y_i<\frac{l_1}{n}\left(1-\sqrt{\frac{l(n-H)}{H(n-l)}}\right) \qquad\text{is at most} \qquad n-l 
 }
\end{equation}

\end{theorem}

\begin{proof}

As detailed in \cite{Jensen1999}, there are two approaches to prove such Laguerre-Samuelson style Inequality.
A first one, following Laguerre's style using Cauchy-Schwarz inequality, whereas a second one, following Samuelson's style \cite{samuelson1968deviant} uses a clustering argument.
We provide both proofs in Appendix section \ref{zerosum-proof}.
The proof is based on the following lemma. 
\begin{lemma}
Let
\[
r :=\#\left\{ i:\; y_i \le \frac{l_1}{n} \left( 1-\sqrt{\frac{l(n-H)}{H(n-l)}} \right) \right\}
  = \#\left\{ i:\; y_i\le \mu-t \right\},
\]
where
\[
 t:=\mu\sqrt{\frac{l(n-H)}{H(n-l)}}.
\]
Then
\begin{equation}
r\le \frac{n\sigma^2}{\sigma^2+t^2}.
\end{equation}
\end{lemma}
\end{proof}

\paragraph{Consequence.}
Theorem~\ref{Jensen} shows that the number $r$ of coefficients removed
by the thresholding rule satisfies
\[
r \leq n-l.
\]
Consequently, the remaining active support satisfies
\[
\nu = n-r \geq l.
\]
This result establishes a rigorous lower bound on the active support
size and provides a direct relationship between the parameter $l$
and the sparsity induced by the projection.
Decreasing $l$ increases the admissible number of removed
coefficients and therefore promotes sparser solutions, whereas
increasing $l$ enforces a larger active support.\\
Although there is generally no closed-form expression relating ($l$) to an exact sparsity percentage since the final support also depends on the distribution of the input coefficients, the monotonic behavior makes the parameter straightforward to calibrate in practice.
\begin{figure}[ht]
    \centering
    \includegraphics[width=0.7\textwidth]{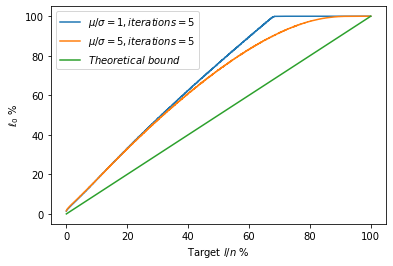}
    \caption{ Synthetic Gaussian distribution with $\frac{\mu}{\sigma} $ active size: 
   Average active-set size after five PLCP iterations as a
function of the target parameter $l$ for synthetic Gaussian data.
Results are reported for different values of $\mu/\sigma$ and
compared with the theoretical bound (green).}
    \label{Bound_low}
\end{figure}
Figure \ref{Bound_low}  shows good agreement between the bound established in Theorem~3.7 and the experimental Gaussian results.
 The theoretical lower bound provided by Theorem~\ref{Jensen}
      guarantees that the active set size is at least $l$ and the relationship between ($l$) and the achieved sparsity.
Consequently, a desired sparsity ratio can be achieved reliably through a simple calibration procedure (for example, a monotonic search on a small validation set), after which the selected value of ($l$) generalizes consistently across the evaluation data.
 Moreover,
 Figure \ref{Bound_low} illustrates the relationship between the target ($l$) and the achieved sparsity and thus the practical calibration procedure used to reach a target sparsity level.

\begin{figure}
    \begin{center}
    \includegraphics[width=0.7\textwidth]{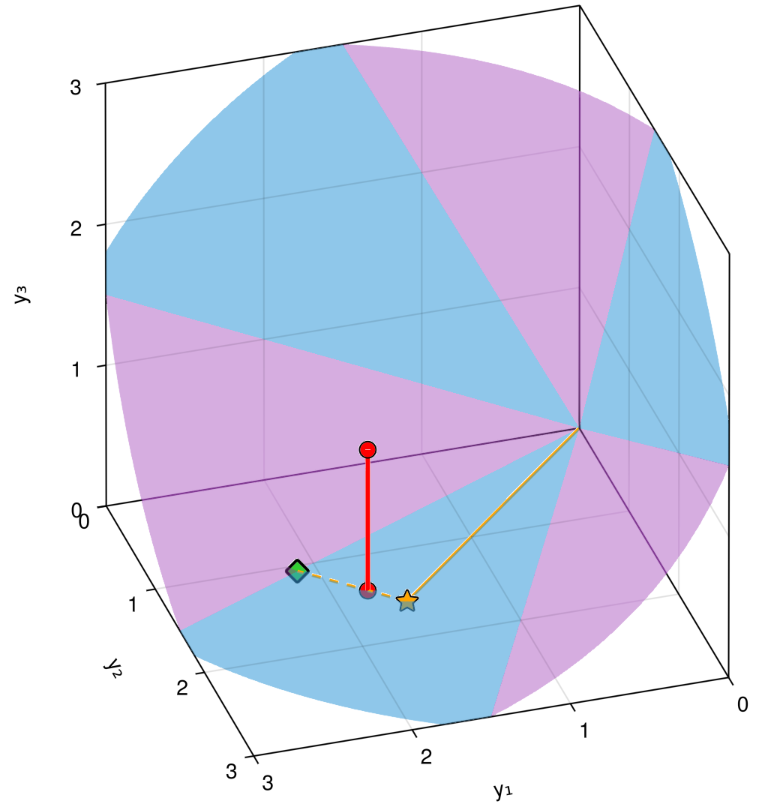}  
    \end{center}
\caption{Illustration of Algorithm~\ref{PLCP} with $l = 1.8$ starting $\mathbf{x} = (1.78, 1.4, 1)$. The boundary of the CAI cone is represented by the blue and orange surfaces. The red circle in the interior of the cone is the starting $\mathbf{x}$. After one iteration of the algorithm, $x_3$ is set to 0 and we obtain the second red dot.
    This point belongs to the blue face in the $x_1$--$x_2$ plane which is itself a two-dimensional CAI cone.
    The yellow star represents $\mathbf{d}$ and the green circle is the output $\lambda \mathbf{x} + (1-\lambda) \mathbf{d}$.}
\end{figure}

\subsection{Convergence of the active set selection}
\label{convergence}

\subsubsection{Finite-time convergence} 
Given a threshold $\alpha \ge 0$, define the hard-thresholding operator $T_\alpha:\mathbb{R}^n_+\to\mathbb{R}^n_+$ by
\begin{equation}
\label{hard_threshold}
\bigl(T_\alpha(\vx)\bigr)_i
=
\begin{cases}
x_i, & \text{if } x_i \ge \alpha,\\
0,   & \text{otherwise},
\end{cases}
\qquad i = 1,\dots,n.
\end{equation}

\begin{proposition}[Finite-time convergence of the active support selection]
 The iterative loop for computing $\alpha$  converges in at most $n$ iterations to a fixed point. More precisely, there exists $K \le n$ such that:

\begin{equation}
\boxed{
\begin{aligned}
\vx^{(K+1)} = \vx^{(K)},\\
\vx^{(K)} = T_{\alpha(\vx^{(K)})}\bigl(\vx^{(K)}\bigr).
\end{aligned}}
\end{equation}
\end{proposition}

This follows immediately from the fact that every decreasing sequence of natural numbers is eventually constant.

\subsubsection{\texorpdfstring{Tightness of the $\mathcal{O}(n)$ Bound}
{Tightness of the O(n) Bound}}

A natural question is whether the bound $K \le n$ of the previous proposition can be improved, for instance to $\mathcal{O}(\log n)$, by exploiting the fact that several entries may fall below the threshold at a single step. The following argument, classical in the robust-statistics literature, shows that no such universal tightening is possible: there exist inputs on which the iteration removes exactly one entry per step. The obstruction is the so-called \emph{masking effect} \cite{davies1993,huber1981}, by which a small group of extreme entries inflates the summary statistics on which the threshold depends and thereby ``masks'' other, less extreme entries that would otherwise be eliminated. The same phenomenon is discussed in \citet[Section~11.1]{huber1981} in the context of outlier-rejection rules, where it is precisely the inability of low-breakdown procedures to peel off more than one outlier per pass that motivates high-breakdown alternatives.

\paragraph{Adversarial construction.}
In our setting the threshold $\alpha(\vx)$ is a smooth function of the $\ell_1$ and $\ell_2$ norms of the current iterate (equivalently, of the CAI-type score $H(\vx)$). Removing an entry produces a finite, controllable change in these norms and hence in $\alpha$. We build an adversarial input recursively, from the stable core outward. Fix a terminal set $S_n$ for which the iteration has already converged (e.g.\ a few equal positive values). Assuming a sequence $S_{k+1}=\{y_{k+1},\dots,y_n\}$ has been constructed so that the iteration peels it one entry at a time, we prepend a single new entry $y_k$ and require
\[
y_k \;<\; \alpha\bigl(\{y_k\}\cup S_{k+1}\bigr) \;\le\; y_{k+1}.
\]
The map $y_k \mapsto \alpha(\{y_k\}\cup S_{k+1}) - y_k$ is continuous, is negative for $y_k$ close enough to $y_{k+1}$ (the entry survives), and is positive for $y_k$ sufficiently small (the entry inflates the norms enough to leave the threshold above it). The Intermediate Value Theorem therefore guarantees the existence of an admissible $y_k$, and recursion produces a sequence on which the iteration removes exactly one entry per step.

\paragraph{Toy illustration.}
The masking mechanism is most transparent for the simpler threshold $\bar y - \alpha\,s(y)$ based on the empirical mean and standard deviation. Take $\vy = \{-100,\,0,\,8,\,10,\,10\}$ with multipliers $\alpha=1$ initially and $\alpha=1.4$ after the first reduction:
\begin{itemize}
\item Iteration 1: $\bar y \approx -14$, $s(y) \approx 43$, threshold $\approx -57$. The extreme value $-100$ is removed; the values $0$ and $8$ are \emph{masked} by the inflated standard deviation.
\item Iteration 2: on $\{0,8,10,10\}$, $\bar y = 7$, $s \approx 4$, threshold $\approx 1.4$. The variance has collapsed and $0$ is now exposed and removed.
\item Iteration 3: on $\{8,10,10\}$, the same collapse exposes $8$, which is removed.
\item Iteration 4: $\{10,10\}$ has zero variance and the iteration terminates.
\end{itemize}
The general staircase construction is the recursive analogue of this $5$-point example.

\paragraph{Consequence.}
The bound $K \le n$ is therefore tight up to constants: no universal sub-linear bound on the number of support-update iterations can hold, because the threshold is computed from internal sample statistics that can be manipulated by an adversarial spacing of the input. The empirical observation (Figure~\ref{active}) that $K$ is small in practice is correspondingly a statement about typical inputs, not a worst-case guarantee, and is consistent with the absence on natural data of the exponentially-spaced staircase that the masking construction requires.

\subsubsection{Comparison with other hard-thresholding iterations}

We compare our hard-thresholding iteration with T\textsc{orrent} \cite{bhatia2015} in detail in Appendix~\ref{sec:bhatia}.
At a high level both methods can be read as instances of a long-standing heuristic, going back to Legendre, that alternates between selecting a subset of indices by thresholding and refitting on that subset. T\textsc{orrent} brings such a selection step inside the regression loop, whereas PLCP repurposes the same mechanism for projecting onto a structured sparsity cone. The two methods differ in what is being fitted (a regressor vs.\ a projection), in how the threshold is set (fixed cardinality vs.\ cone-induced), and in the type of guarantee obtained (geometric statistical recovery vs.\ deterministic finite-time termination).

Viewed jointly, by Luo \cite{luo1992} and our analysis (compared in detail in Appendix~\ref{sec:luo1992}) describe two complementary regimes for iterative methods on LCP-like problems: a coordinate-wise regime with linear convergence under mild structural assumptions, and a block-thresholding regime where the combinatorial structure of the active support directly delivers finite-time termination. The PLCP iteration sits in the latter regime, which is what makes its complexity $\mathcal{O}(Kn)$ with a very small $K$ in practice.

See Appendix~\ref{sec:rousseeuw1984} and Appendix~\ref{sec:huber1981} for the detailed comparison with \cite{rousseeuw1984} and \cite{huber1981} respectively.

\subsection{Complexity Analysis of PLCP}

The iterative PLCP
algorithm typically converges in only a few support updates. Denoting by $K$
the number of thresholding iterations, its actual computational cost is
$\mathcal{O}(Kn)$ rather than $\mathcal{O}(n^2)$. When $K$ remains small, the
algorithm exhibits near-linear behavior in practice.\\
An alternative is a sorted implementation which avoids the quadratic iterative thresholding loop by sorting the coefficients once and scanning the possible active support sizes using prefix sums. 
The total complexity is therefore the complexity of the sort:
$
T(n)
=
\mathcal{O}(n\log n)
$.
 The sorted
formulation systematically incurs the full sorting cost
$\mathcal{O}(n\log n)$.
Moreover, in most sparse optimization problems, the final active support size
$\nu$ is typically much smaller than the ambient dimension $n$
($\nu \ll n$). In such settings, sorting the entire vector introduces a
significant amount of unnecessary computation.
Rather than sorting all coefficients, the Topk method progressively
extracts only the $k$ largest entries and tests whether the resulting support
satisfies the PLCP fixed-point condition. 
The resulting complexity for the Topk method is
\begin{equation}
\mathcal{O}(n + k_{\max}\log k_{\max}),
\end{equation}
which improves over the full sorting cost $\mathcal{O}(n\log n)$ whenever \(k_{\max} \ll n\).

\subsection{Related Work and Positioning of PLCP and PECP }
\subsubsection{Positioning versus SOCP methods}
Conic programming provides a unified framework for convex optimization problems \cite{Boyd-convex,SOC2,SOC1}. Most existing methods involving second-order cone programming (SOCP) rely on two major classes of approaches.

\begin{itemize}

\item \textbf{Interior-point methods.}  
These approaches exploit the geometry of the cone through barrier functions,
typically of the form $-\log\left(x_0^2-\|\vx\|_2^2\right)$
and follow a central path toward the optimal solution. Although highly
effective, the resulting trajectory is not based on an explicit affine
parameterization.

\item \textbf{Projection-based methods.}  
Many optimization algorithms, such as projected gradient methods, proximal
algorithms, ADMM, and alternating projections, rely on Euclidean projections
onto the cone.
These methods use the cone geometry implicitly through projection operators,
but do not introduce an explicit parameterized trajectory.
\end{itemize}
In contrast to standard SOCP projection approaches, PLCP explicitly
constructs an affine parametric path toward the CAI boundary.
PLCP and PECP may therefore be interpreted as a geometric parametric alternative to
standard SOCP projections, combining analytical simplicity with additional
flexibility while preserving the same sparsity properties.
\subsubsection{Positioning PLCP and PECP versus GSP method}

The GSP constraint (Group Sparse Projection), following the definition in \cite{GSP}, is given by
\[
    GSP(\vx)=\left(\sum_{i=1}^r \frac{\sqrt{n_i}-\|x_i\|_1}{\sqrt{n_i}-1}\right).
\]

GSP addresses the projection problem through a scalar dual root-finding procedure derived from optimality conditions. Each iteration has linear complexity in the problem dimension, and the overall complexity is reported as $\mathcal{O}(n \log(1/\varepsilon))$, where $\varepsilon$ denotes the target accuracy. A bisection fallback ensures robustness, with worst-case complexity bounded by $\mathcal{O}(n \log n)$.

Note that the GSP-Hybrid constraint is non-convex, whereas PLCP
exploits the convex geometry of the CAI interior in the
high-sparsity regime.
The active set selection is computed directly in the primal space via a finite sequence of active-set updates combined with thresholding operations, yielding a complexity of $\mathcal{O}(kn)$, where $k$ denotes the number of active-set refinements.
By combining closed-form projections, conic geometry, and
hardware-friendly structured sparsity, the proposed framework offers
a scalable foundation for efficient inference and training of modern
neural networks.

\subsection{ Bilevel PECP Projection}

Recently, bilevel optimization (BLO) \cite{overview,Bilevel-pruning,bilevel-optim,BLO} has become popular because it is a powerful method in modeling problems that involve optimizing nested objective functions.
BLO is an optimization problem that involves
two levels of hierarchy (i.e., upper and lower levels), wherein
obtaining the solution to the upper-level problem requires
solving the lower-level one.  
In this paper, we propose a specific bilevel method for enforcing structured sparsity.
We provide a single-level characterization of the proposed bilevel operator as follows.\\
Let \(Y\in\mathbb{R}^{n\times n}\) with columns
\(y_1,\dots,y_n\in\mathbb{R}^n\). Define the vector of column \(\ell_2\) norms
\[\vv \;:=\; (\|y_1\|_2,\dots,\|y_n\|_2)\in\mathbb{R}^n.\]
Define the specific bilevel optimization problem and the bilevel operator \(BP_{PECP,2}^l(Y)\) as follows:
\begin{equation}
\begin{aligned}
\hat{u} = \arg \min_{\vu} \|\vu - \vv\|_2 \;\;\text{s.t. } H(\vu) < l.\\
x_j = P_2^{\hat u_j}(y_j) \quad \forall j\in\{1,\dots,n\}.
\end{aligned}
\label{eq:def_bilevel_x}
\end{equation}
The resulting bilevel PECP  algorithm is computationally inexpensive:
\begin{algorithm}
\begin{algorithmic}
\caption{bilevel PECP projection, ($BP^{PECP,2}_l(\mY)$)}
\label{algo:li12proj}
\State \textbf{Input:} $\mY \in \sR^{n\times n}$, $ l > 0$
\State $\hat{\vu} \gets P^{PECP}_l\bigl(\|\vy_1\|_2,\dots,\|\vy_n\|_2\bigr)$
\State $ x_j = P_2^{\hat u_j}(y_j)
\quad\leftrightarrow\quad
\begin{cases}
y_j, & \text{if } \|y_j\|_2 \le u_j,\\[4pt]
\dfrac{u_j}{\|y_j\|_2}\,y_j, & \text{if } \|y_j\|_2 > u_j.
\end{cases}  $
\State \textbf{Output:} $\mX = [\vx_1,\dots,\vx_n]$
\end{algorithmic}
\end{algorithm}

\begin{proposition}
Minimizing the Frobenius distance in matrix space is
equivalent to minimizing the Euclidean distance between the vectors
of column norms:
\[
\min_X\frac12\|Y-X\|_F^2
=
\min_{u:\,H(u)\leq l}
\frac12\|v-u\|_2^2.
\]
Since $\widehat u=P_l^{\mathrm{PECP}}(v)$ is the Euclidean CAI
projection of $v$, the matrix obtained from
$x_j=P_{\widehat u_j}^{2}(y_j)$ is the corresponding exact
Frobenius projection.

See proof in appendix \ref{PECP-Frob}
\end{proposition}

\section{Experimental results}

\subsection{Benchmark of PLCP} 
We compare our method in terms of computation time with the Group Sparse LASSO using the GSP method \cite{GSP}. 
For the implementation of the GSP-Hybrid method, we use the PyTorch implementation provided by Ohib et al.  
\footnote{ \url{https://github.com/riohib/gsp-for-deeplearning} }    
We compare PLCP with a structured row/column pruning baseline inspired by the structured pruning method in \cite{structured}. For each attention matrix, rows and columns are ranked according to their $\ell_2$ norm and the least important structures are removed while preserving the CLS row and column.
Our PyTorch implementation is provided in the supplementary material.

\begin{figure}[ht]
    \centering
    \includegraphics[width=0.7\textwidth]{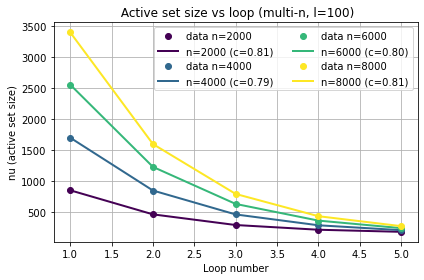}
    \caption{ Convergence of the active support size for different size of the Gaussian data  as a function of loop  number. }
    \label{active}
\end{figure}

The PLCP iteration can be interpreted as an adaptive thresholding scheme, where the threshold is updated from the current active coefficients. 
Under the assumption that the threshold dynamics is locally contractive near its fixed point and that the tail of the coefficient distribution is sufficiently smooth, the active-set size is expected to converge exponentially fast toward a limiting value $\nu_\infty$. 
\[
\nu_k - \nu_\infty \approx (\nu_0-\nu_\infty) e^{-ck},
\]
where $c>0$ is an effective decay constant. 
Figure \ref{active} shows the convergence of the active support size for different size of the Gaussian data as a function of loop  number.
The best exponential fit is with parameter $c\approx 0.8$.

\paragraph{Probabilistic geometric convergence of the active-set selection.}
We prove theoretically finite termination, but exponential convergence is only empirically observed.
However, a geometric convergence rate can be established under a probabilistic
assumption on the distribution of the active coefficients.

As shown with inequality~\ref{eq:zscore}, the threshold is a clipping on the z-score of the samples only depending on $l$ and $n$. Therefore, the Cumulative Density Function (CDF) of the distribution of the samples, evaluated at this threshold, gives a probability $p_i$ for a sample to remain in the active set at iteration $i$.
This matches a geometric convergence rate if $p_i$ were constant which we know is not the case here as we don't expect the active set size to converge to zero.
The complex task of analysing this series is well studied as we discuss in appendix section~\ref{an:lit}.

Instead, assume that, conditionally on the current active set, each incorrectly
active coefficient has a probability at least \(p>0\) of falling below
the updated threshold and being removed at the next iteration.
Let
\(S_k\) denote the active support at iteration \(k\), let \(S_\star\) be
the final support, and define
\[
e_k = |S_k \setminus S_\star|.
\]
Then
\[
\mathbb{E}\!\left[e_{k+1}\mid e_k\right]
\leq (1-p)e_k.
\]
By iterating this inequality, we obtain
\[
\mathbb{E}[e_k]
\leq (1-p)^k e_0.
\]
Hence, the expected number of incorrectly active coefficients decreases
geometrically. In particular, the expected active set converges toward
the limiting support at an exponential rate with contraction factor
\(1-p\).

\begin{figure}[ht]
   \centering
   \includegraphics[width=0.6\textwidth]{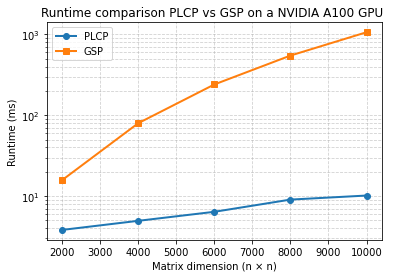}
   \includegraphics[width=0.6\textwidth]{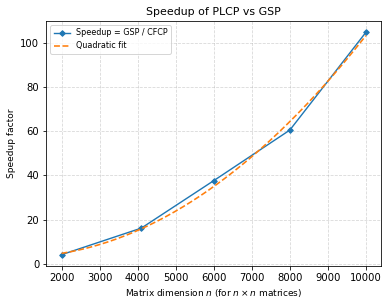}
   \caption{Up: runtime comparison of PLCP vs GSP as a function of the size of matrices.\\
bottom: quadratic fit of the speed-up of PLCP vs  GSP as a function of the size of matrices using an NVIDIA A100 GPU.}
   \label{PLCP-Prune-time}
\end{figure}

The key difference between PLCP and GSP is dimensional:
GSP operates on a vectorized representation of the matrix, involving $n^2$ variables, and requires multiple passes over all entries due to its root-finding procedure.
In contrast, PLCP exploits the group structure by reducing the problem to $n$ variables via column norms, performing the core projection in $\mathcal{O}(n\log n)$ similar to the pruning method.
This structural difference explains the observed speed-up.
Note that the GSP-Hybrid method constraint is non-convex, whereas PLCP enforces the convex geometry of the CAI interior in the high sparsity regime.

\begin{figure}[ht]
   \centering
   \includegraphics[width=0.7\textwidth]{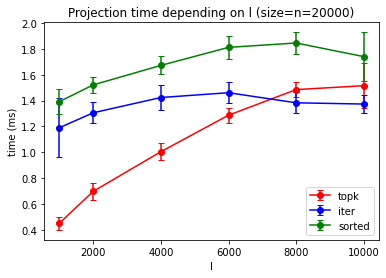}
   \caption{Runtime comparison of different one-dimensional PLCP implementations: iterative PLCP (blue), GPU-friendly TopK PLCP
(red), and fully sorted PLCP (green) on a M3 apple CPU }
   \label{PLCP-time}
\end{figure}

Figure~\ref{PLCP-time} shows that the runtime of iterative PLCP is always better than that of PLCP with exhaustive sorting, while iterative PLCP is better than PLCP-TopK for large datasets.
The runtime of iterative PLCP is very stable with respect to the dimension, whereas that of PLCP-TopK is not. 
Note that structured pruning exhibits a runtime comparable to
that of the GPU-friendly TopK PLCP implementation.

\subsection{Sparsification of Attention Matrices in Transformer Architectures}

\paragraph{Constraint optimization }
Let $\mW \in \mathbb{R}^{n \times n}$ denote the attention matrix, where $n$ is the number of tokens. 
Let $\vz $ represent the ground-truth labels, and $\vz^*$ denote the class probabilities predicted by the softmax classifier.
To sparsify the weights $\mW$ of the attention matrix, we employ the bilevel projection method $BP^{PLCP,2}$ as a constraint to enforce structured sparsity in the model. The global optimization criterion is defined as:
\begin{equation}
 \label{crit}
 \min_{\mW}  
 \quad \phi(\vz,\vz^*) 
 \quad \text{s. t.} \quad CAI(\mW) \le l,
\end{equation}
where $\phi(\vz,\vz^*)$ is the cross-entropy loss.
To minimize this criterion, we follow the work developed by \cite{double} and \cite{frankle-init} who proposed a double-descent masked-gradient algorithm.
We replace their threshold with our bilevel projection.\\
\paragraph{We apply our closed-form conic projection to the sparsification
of attention matrices in Transformer architectures \cite{Transformer}.}
Specifically, we compare our learned diagonal mask, obtained via bilevel projection, against the uniform diagonal mask of BigBird \cite{Bigbird}.
The classification framework is implemented in PyTorch, including the model, schedulers, and loss functions. We chose the Adam optimizer (Adaptive Moment Estimation) \cite{Adam} and  the smooth SiLU activation function.
For all sparsity levels and all datasets, we set the number of training epochs to 15, the batch size to 64 and the learning rate to 2e-5.
We plot the curves of accuracy as a function of sparsity. 
The numerical labels in the following figures denote the window
size for BigBird and the constraint level $l$ for PLCP, GSP,
and structured pruning, respectively.

\paragraph{Experiment on a natural language processing (NLP) task from the GLUE dataset \cite{GLUE}}
While pretrained models such as BERT are fully dense Transformers, some later architectures (e.g., BigBird, Longformer) introduce sparse attention mechanisms. 
Specifically, we compare these methods with our PLCP projection on a pretrained transformer-based model \cite{BERT_I}. For the QNLI, QQP  and MNLI datasets, we used attention matrices with 256 tokens, and 128 tokens for the SST-2 dataset.
Recall that a vertical band in an attention matrix corresponds to a \emph{global token} attending to all tokens, whereas a horizontal band corresponds to all tokens attending to a global token.
Such global attention patterns are known to play a critical role in Transformer-based classification models, as they enable long-range information aggregation and act as routing hubs for semantic context.
From a theoretical perspective, removing these global interactions may severely limit the expressive power of the self-attention mechanism, since the first token (e.g., \texttt{[CLS]}) often serves as a summary representation of the entire sequence.
Consequently, preserving the first row and the first column of the attention matrix  ensures that global pairwise interactions are maintained, even under strong sparse constraints.
Therefore, in all subsequent experiments, we explicitly include for the PLCP, GSP, Pruning and BigBird masks the first row and the first column in the  sparsity mask.\\
In order to capture local dependencies, it is natural to focus on the diagonals of the attention matrix. The diagonal transformation used in PLCP reshapes the attention matrix into a structured representation where diagonal interactions become axis-aligned, enabling structured sparsity projections. 

\textbf{The SST-2 dataset } The Stanford Sentiment Treebank consists of sentences from movie reviews and human annotations of their sentiment. The task is to predict the sentiment of a given sentence \cite{socher}. This dataset is composed of 67,349 samples and is available in the GLUE dataset.

\begin{figure}[ht]
\centering
\includegraphics[width=0.49\textwidth]{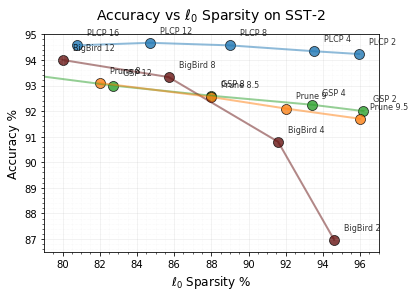}
\caption{ PLCP vs GSP, Pruning  and BigBird: sparsity–accuracy trade-off.  SST-2 dataset }
\label{SST2}
\end{figure}

\textbf{The Question Natural Language Inference (QNLI)} dataset  \cite{QNLI} is derived from the Stanford Question Answering Dataset (SQuAD) and reformulates question answering as a binary sentence-pair classification task.
The task is to determine whether the context sentence contains the answer to the question.
This dataset contains approximately 100K labeled examples.

\begin{figure}[ht]
\centering
\includegraphics[width=0.49\textwidth]{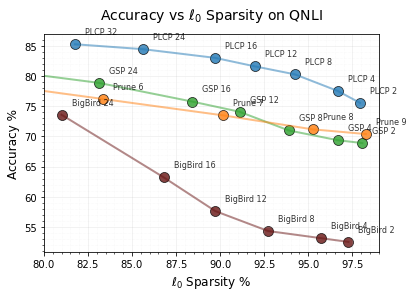}
\includegraphics[width=0.49\textwidth]{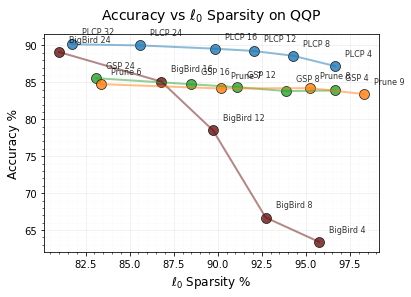}
\caption{ PLCP vs GSP, Pruning and BigBird: sparsity–accuracy trade-off. Left: QNLI dataset. Right: QQP dataset.}
\label{QNLI}
\end{figure}

\textbf{The Quora Question Pairs dataset (QQP)} is a collection of question pairs from the community
question-answering website Quora. The task is to determine whether a pair of questions are semantically equivalent. This dataset consists of $\approx$ 400 K labeled samples.\\

\textbf{The Multi-Genre Natural Language Inference Corpus} \cite{MNLI} is a crowdsourced
collection of sentence pairs with textual entailment annotations. Given a premise sentence
and a hypothesis sentence, the task is to predict whether the premise entails the hypothesis (entailment),
contradicts the hypothesis (contradiction), or neither (neutral). We report accuracy between true labels and predicted labels. This dataset consists of $\approx$ 400 K samples with labeled  classification.

\textbf{WiC (Word-in-Context)} \cite{WIC} is a word sense disambiguation task
cast as binary classification of sentence pairs. Given two text snippets and a polysemous word that
appears in both sentences, the task is to determine whether the word is used with the same sense in
both sentences. This dataset consists of 6,000 samples with labeled  classification.

\begin{figure}[ht]
\centering
\includegraphics[width=0.49\textwidth]{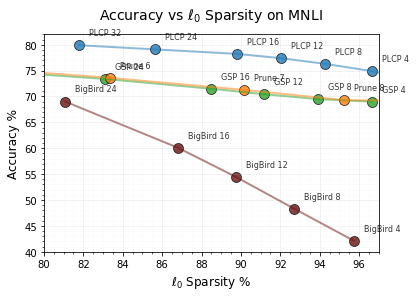}
\includegraphics[width=0.49\textwidth]{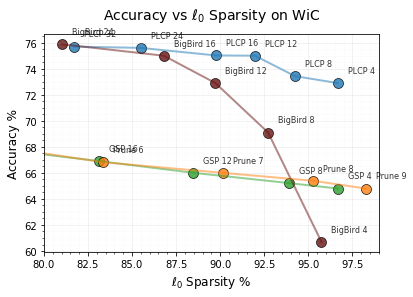}
\caption{ PLCP vs GSP, Pruning and BigBird: sparsity–accuracy trade-off. Left: GLUE  MNLI dataset; Right: SuperGLUE  WiC dataset }
\label{MNLI}
\end{figure}

Figures \ref{SST2}, \ref{QNLI} and \ref{MNLI} show accuracy-sparsity curves across five datasets. 
The accuracy-sparsity curves of PLCP, GSP, and structured pruning follow similar overall trends, but PLCP consistently dominates the competing methods. It delivers an accuracy gain of about $2\%$ on the relatively easy SST-2 benchmark, while the margin widens to up to $6\%$ on larger and more challenging datasets such as QNLI, QQP, and MNLI. On the particularly difficult WiC benchmark, PLCP achieves gains of up to $10\%$, highlighting its ability to preserve discriminative information even under aggressive sparsification.\\
PLCP leverages a column-wise bilevel projection that induces highly structured sparsity aligned with diagonal patterns while preserving locality after reshaping. This enables the model to retain the most informative interactions even at high sparsity levels.\\
In contrast, Group Lasso regularization promotes coarse structured sparsity by selecting or removing entire parameter groups. Although effective at the group level, such masks lack the flexibility required to capture fine-grained local dependencies, particularly diagonal structures, since sparsity is enforced uniformly within each group.\\
Structured pruning is a coarse-grained ranking strategy that enforces row-wise and column-wise sparsity. Although efficient, it is inherently unable to capture diagonal structures and fine-grained local dependencies, which often carry the most informative interactions in attention matrices.\\
 The fixed BigBird mask performs adequately in the low-sparsity regime (large masks), but its rigid predefined structure becomes a significant limitation at high sparsity levels, where adaptivity becomes crucial.\\
These fundamental structural differences explain the consistently superior accuracy-sparsity trade-off achieved by PLCP compared with GSP, structured pruning and BigBird across all datasets.

\paragraph{Accuracy Stability Across Random Seeds}

In addition to reporting the mean classification accuracy, we evaluate
the robustness of each sparsification method with respect to the random
seed. 
Our experiments involve four methods, five datasets, six
hyperparameter settings, and three random seeds.

For each method $m$, dataset $d$, and hyperparameter setting $h$,
we compute the standard deviation of the classification accuracy
across the three experimental seeds.

\[
\sigma_{m,d,h}
=
\operatorname{Std}_{s}
\left(
\operatorname{Accuracy}_{m,d,h,s}
\right),
\]
where $s$ denotes the random seed.

In our experiments, several independent runs are considered. Therefore, $\sigma_{m,d,h}$ measures the
variability of the accuracy between these runs for a fixed method,
dataset, and hyperparameter value.
Second, to obtain a single global stability measure for each method, we
average these standard deviations over all datasets and all considered
hyperparameter values.
Thus, the global stability criterion can be written as
\[
\boxed{
\overline{\sigma}_m
=
\operatorname{Mean}_{d,h}
\left[
\operatorname{Std}_{s}
\left(
\operatorname{Accuracy}_{m,d,h,s}
\right)
\right].
}
\]

The random
seeds are first used to estimate the variability of each individual
experimental configuration. These configuration-wise standard deviations
are then averaged across datasets and hyperparameter values.
Consequently, $\overline{\sigma}_m$ provides a compact measure of the
overall sensitivity of method $m$ to random initialization and other
seed-dependent effects. A lower value of $\overline{\sigma}_m$ indicates
that the method produces more consistent classification accuracy across
independent experimental runs.

We report the standard deviation of accuracy in Table~\ref{STD}, which allows us to compare
the overall stability of the four sparsification strategies considered
in this work: BigBird, PLCP, GSP, and structured pruning, independently
of their average accuracy and sparsity levels.
To compute this average, the number of runs is 35 for PLCP and GSP, whereas it is 30 for BigBird and Pruning. Note that BigBird was originally proposed for large attention matrices. Overall, the four methods exhibit comparable stability.

\begin{table}[ht]
\centering
\begin{tabular}{|c|c|c|c|c|c|}
\hline
 & BigBird  & PLCP & GSP & Pruning \\
\hline
Accuracy STD ($ \% $) & 0.2460 & 0.2531 & 0.24449 &0.2147\\
\hline
\end{tabular}
\caption{Average standard deviation of classification accuracy across
random seeds for BigBird, PLCP, GSP, and structured pruning.} 
\label{STD}
\end{table}

\begin{table}[ht]
\centering
\begin{tabular}{|c|c|c|c|c|c|c|c|}
\hline
$l$ & 2 & 4 & 8 & 12 & 16 & 24 & 32 \\
\hline
PLCP & 75.54 & 77.27 & 80.37 & 82.03 & 83.17 & 84.63 & 85.49 \\
\hline
PECP & 75.59 & 77.12 & 80.18 & 81.86 & 83.00 & 84.46 & 85.48 \\
\hline
Difference & 0.05 & 0.15 & 0.19 & 0.17 & 0.13 & 0.17 & 0.01 \\
\hline
\end{tabular}
\caption{Comparison between PLCP and PECP on QNLI dataset for different CAI constraint levels $l$.}
\label{pecpplcp}
\end{table}

Table \ref{pecpplcp} shows that PLCP and PECP achieve nearly identical
classification accuracy on QNLI over the entire range of
constraint levels. The mean absolute difference across all values of $l$ is only
$0.12\%$. This empirical result is consistent with our theoretical
analysis.

\paragraph{Ablation study}
To better understand the impact of the PLCP projection, we conduct a comprehensive ablation study.  We consider the following variants: Baseline, BigBird , PLCP, GSP and Pruning.
All variants are evaluated under identical training protocols, initialization, and hyperparameters.

\begin{table}[ht]
\centering
\begin{tabular}{|c|c|c|c|c|c|}
\hline
Accuracy&  Baseline & BigBird  & PLCP & GSP & Pruning \\
\hline
SST-2  & 92.7 & 88 & 94.56 & 92.43 & 92.80\\
\hline
QNLI & 88.52 & 57  & 82.95 & 74.05 & 73.49\\
\hline
QQP & 88.65 & 78.46 & 89.66 &84.31 & 84.17\\
\hline
MNLI & 78.9  & 54.23   & 78.20 & 70.51 &71.19\\
\hline
WiC & 69.55  & 72.76  & 75.13 & 66.15 & 65.16\\
\hline
\end{tabular}
\caption{Ablation. Accuracy at $90 \%$ attention sparsity. Baseline (no projection),  BigBird, PLCP, GSP and Pruning  }
\label{Ablation}
\end{table}

Table~\ref{Ablation} reports the impact of each ablation setting on the sparsity level and predictive accuracy. 
Notably, on the SST-2 dataset, enforcing $90\%$ sparsity leads to an improvement in accuracy, as it effectively removes noisy tokens.\\
Table~\ref{Ablation} shows that QQP and WiC are highly robust to an attention sparsity level of 90\%, whereas QNLI and MNLI exhibit a stronger dependence on dense attention. 
The larger performance degradation observed on QNLI at 90\% sparsity may be explained by the nature of the task. QNLI relies on semantic interactions between the question and the corresponding sentence, which may require a more distributed attention pattern. At very high sparsity levels, removing a large fraction of token-to-token interactions may therefore discard relevant cross-token dependencies. 
MNLI is a challenging natural language inference task, for which removing 90\% of the token-to-token interactions may naturally have a more detrimental impact on performance.
GSP and Pruning achieve comparable performance, while PLCP consistently outperforms both methods by approximately 5 to 9 percentage points. BigBird performs substantially worse than PLCP, GSP, and Pruning on most datasets, with WiC being the notable exception.

\section{Discussion and future work}
\subsection{Linguistic analysis}
We analyze the inference computational cost of the proposed PLCP in section \ref{inference} using a BERT-like attention block \cite{BERT_I}. 
An important direction for future work is to investigate the linguistic impact of aggressive attention sparsification. Previous studies have shown that some attention heads capture linguistically interpretable patterns and that such specialized heads tend to be preserved during pruning \cite{voita}. In our experiments, we qualitatively observe that sparsification tends to suppress attention associated with punctuation and some less informative tokens while preserving the overall semantic content. A systematic linguistic analysis would therefore be valuable to determine whether PLCP preferentially preserves syntactic and semantic dependencies as sparsity increases. In particular, future work could compare PLCP with BigBird and structured pruning at matched sparsity levels by measuring the semantic content of the attention patterns that are retained.\\
\subsection{Extension to Mixture-of-Experts Architectures }
Although this work focuses on structured sparsification of Transformer attention matrices, the proposed PLCP framework naturally extends to a much broader class of structured matrix optimization problems. In particular, \emph{Mixture-of-Experts} (MoE) architectures \cite{MOE1,MOE2,MOE3} constitute a promising application domain.

In a standard Transformer, the feed-forward network (FFN) consists of two dense matrices
\[
W_1\in\mathbb{R}^{d\times4d},
\qquad
W_2\in\mathbb{R}^{4d\times d},
\]
shared by all input tokens.

In contrast, an MoE layer replaces this single FFN by a collection of $E$ independent experts,
\[
\left\{W_1^{(e)},W_2^{(e)}\right\}_{e=1}^{E},
\]
where each expert possesses its own pair of weight matrices. A lightweight routing network selects only the most relevant experts for each input token, typically one expert (Switch Transformer) or two experts (Mixtral), considerably reducing the computational cost while maintaining a very large overall model capacity.

From a matrix viewpoint, the collection of expert weights can be interpreted as a block-diagonal operator,
revealing a highly structured matrix organization that is particularly attractive for structured sparsification.
Each expert matrix can be projected independently using PLCP or PECP, naturally producing structured sparsity patterns such as row-wise, column-wise or diagonal sparsity. 
This could reduce both memory consumption and inference latency
without modifying the routing mechanism.

\section{Conclusion}

In this paper, we introduced the Cone Alignment Index (CAI), a
scale-invariant sparsity measure associated with a Lorentz-cone
geometry.
Building on this formulation, we derived two complementary projections: the \emph{Parametric Linear Conic Projection} (PLCP), an efficient closed-form approximation, and the \emph{Parametric Euclidean Conic Projection} (PECP), the exact Euclidean projection onto the cone boundary. We proved that PLCP and PECP are collinear, differ only by a positive radial scaling factor, and induce exactly the same sparsity threshold, thereby preserving the same active support.

Experiments demonstrate that our approach achieves up to $90\%$ structured attention sparsity on GLUE benchmarks while preserving predictive performance, consistently outperforming GSP, structured pruning, and the sparse-attention BigBird.

By combining closed-form projections, conic geometry, and
structured sparsity, the proposed framework offers a scalable
foundation for efficient inference and training of modern neural
networks.

Future work will extend this framework well beyond attention mechanisms, investigating structured sparsification of general weight matrices and Mixture-of-Experts architectures.

\bibliography{tmlr}
\bibliographystyle{plain}

\newpage
\appendix
\section{Appendix}

\subsection{Asymptotic Distribution of the CAI score}
\label{distribution}

\begin{theorem}[Asymptotic distribution of the CAI score]
\label{thm:hoyer_general}

Let $Y_1,\ldots,Y_n$ be i.i.d.\ random variables such that
\[
0<\mu_2=\mathbb{E}[Y^2]<\infty,
\qquad
\mathbb{E}[Y^4]<\infty,
\]
and define
\[
H_n
=
\frac{\left(\sum_{i=1}^n |Y_i|\right)^2}
{\sum_{i=1}^n Y_i^2}.
\]
Let
\[
\mu_1=\mathbb{E}|Y|,
\qquad
\mu_3=\mathbb{E}|Y|^3,
\qquad
\mu_4=\mathbb{E}[Y^4],
\]
and
\[
m=\frac{\mu_1^2}{\mu_2}.
\]

Then
\[
\frac{H_n}{n}
\xrightarrow{\mathbb P}
m.
\]

Moreover,
\[
\sqrt{n}
\left(
\frac{H_n}{n}-m
\right)
\xrightarrow{d}
\mathcal N(0,\sigma_H^2),
\]
where
\[
\sigma_H^2
=
\nabla g(\mu_1,\mu_2)^\top
\Sigma
\nabla g(\mu_1,\mu_2),
\]
with
\[
g(a,b)=\frac{a^2}{b},
\qquad
\nabla g(\mu_1,\mu_2)
=
\begin{pmatrix}
2\mu_1/\mu_2\\[1mm]
-\mu_1^2/\mu_2^2
\end{pmatrix},
\]
and
\[
\Sigma=
\begin{pmatrix}
\mu_2-\mu_1^2
&
\mu_3-\mu_1\mu_2
\\
\mu_3-\mu_1\mu_2
&
\mu_4-\mu_2^2
\end{pmatrix}.
\]

Equivalently, for large $n$,
\[
\boxed{
\frac{H_n}{n}
\approx
\mathcal N
\left(
m,\frac{\sigma_H^2}{n}
\right).
}
\]
\end{theorem}

\begin{proof}
Define
\[
A_n=\frac1n\sum_{i=1}^n |Y_i|,
\qquad
B_n=\frac1n\sum_{i=1}^n Y_i^2.
\]
Then
\[
\frac{H_n}{n}
=
\frac{A_n^2}{B_n}
=
g(A_n,B_n).
\]

By the law of large numbers,
\[
(A_n,B_n)
\xrightarrow{\mathbb P}
(\mu_1,\mu_2),
\]
and therefore, by continuity of $g$,
\[
\frac{H_n}{n}
\xrightarrow{\mathbb P}
\frac{\mu_1^2}{\mu_2}
=m.
\]

Since $\mathbb{E}[Y^4]<\infty$, the multivariate central limit
theorem gives
\[
\sqrt n
\left[
\begin{pmatrix}
A_n\\ B_n
\end{pmatrix}
-
\begin{pmatrix}
\mu_1\\ \mu_2
\end{pmatrix}
\right]
\xrightarrow{d}
\mathcal N(0,\Sigma).
\]
Applying the delta method to $g(a,b)=a^2/b$ yields
\[
\sqrt n
\left(
\frac{H_n}{n}-m
\right)
\xrightarrow{d}
\mathcal N(0,\sigma_H^2),
\]
which proves the result.
\end{proof}

\begin{corollary}[Gaussian case]
\label{cor:hoyer_gaussian}

If
\[
Y_i\stackrel{\mathrm{i.i.d.}}{\sim}\mathcal N(0,1),
\]
then
\[
\mu_1=\sqrt{\frac{2}{\pi}},
\qquad
\mu_2=1,
\qquad
\mu_3=2\sqrt{\frac{2}{\pi}},
\qquad
\mu_4=3.
\]
Hence
\[
m=\frac{\mu_1^2}{\mu_2}
=\frac{2}{\pi},
\]
and
\[
\sigma_H^2
=
\frac{8}{\pi}
\left(1-\frac{3}{\pi}\right).
\]

The linear-in-$n$ parametrization naturally follows from the relation between
the CAI parameter $l$ and the normalized sparsity. Indeed, for a vector of
dimension $n$, the CAI parameter satisfies $1\leq l\leq n$, and the normalized
sparsity level is approximately characterized by the dimensionless ratio
\[
s \simeq 1-\frac{l}{n}.
\]
Keeping a prescribed normalized sparsity level as the dimension
increases requires $l$ to scale linearly with $n$. This motivates the use of
$l/n$ as the natural dimension-independent parameter for comparing sparsity
regimes across different dimensions.
Therefore the normalized distribution is given by:
\[
\boxed{
\frac{H_n}{n}
\approx
\mathcal N
\left(
\frac{2}{\pi},
\frac{8}{\pi n}
\left(1-\frac{3}{\pi}\right)
\right).
}
\]

\end{corollary}

\subsection{Second-order Gaussian approximation}
\label{second}

Note that the random variable $H(Y)$ where $Y \sim N(0, 1)$ can be equivalently defined as the L1 norm $\|Z\|_1^2$ of a vector $Z$ sampled uniformly over the unit sphere.
This relates our result to \cite[Theorem~3.6]{kabluchko2016meanwidthregularpolytopes} and the particular case for $n = 1$ of \cite[Theorem~1.1]{paouris2012centrallimittheoremprojections} which studies $\|Z\|_1$.\\
Since Theorem~2.2 considers the square of $\|Z\|_1$, our result
can be derived from theirs by applying the delta method.\\
For Gaussian coefficients, the general second-order correction becomes
\[
B_H
=
1-\frac{2}{\pi}.
\]

Hence,
\[
\boxed{
\mathbb E[H_n]
=
\frac{2n}{\pi}
+
1-\frac{2}{\pi}
+
o(1).
}
\]

Interestingly, in the Gaussian case this expression is not merely
asymptotic. Spherical symmetry gives the exact finite-dimensional
identity
\[
\boxed{
\mathbb E[H_n]
=
1+\frac{2(n-1)}{\pi}
=
\frac{2n}{\pi}
+
1-\frac{2}{\pi}.
}
\]

We therefore obtain the refined approximation
\[
\boxed{
H_n
\approx
\mathcal N
\left(
\frac{2n}{\pi}
+
1-\frac{2}{\pi},
\,
\frac{8n(\pi-3)}{\pi^2}
\right).
}
\]

Equivalently,
\[
\boxed{
\mathbb P(H_n\leq l)
\approx
\Phi
\left(
\frac{
\pi l-2n-\pi+2
}{
\sqrt{8n(\pi-3)}
}
\right).
}
\]

Moreover, the asymptotic standard deviation of $H_n$ is
\[
\operatorname{Std}(H_n)
\sim
\sqrt{
\frac{8n(\pi-3)}{\pi^2}
}
\simeq
0.339\,\sqrt{n}.
\]

Moreover, the natural fluctuations of $H_n$ are of order
$\mathcal{O}(\sqrt{n})$. In particular,
\[
\frac{
1-\frac{2}{\pi}
}{
\sqrt{\frac{8n(\pi-3)}{\pi^2}}
}
=
\mathcal{O}\left(\frac{1}{\sqrt{n}}\right).
\]

\subsection{PLCP and PECP share a Common Active-Set Selection}
\label{Euclide}

We now introduce a unified one-sided formulation of PLCP and PECP.
The key observation is that both methods can share the same active-set
selection procedure and therefore the same threshold $\alpha$.
They differ only in the final geometric projection applied once the
active support has been identified.

More precisely, the algorithm is decomposed into two successive stages:
\[
\boxed{
\text{Active-set selection}
\quad\longrightarrow\quad
\text{Geometric projection}.
}
\]

The first stage determines the active support and is common to both PLCP
and PECP. The second stage is method-dependent: PLCP uses an affine
extrapolation parameterized by $\lambda$, whereas PECP performs the
Euclidean projection onto the CAI boundary, parameterized by $t$ or,
equivalently, by the radial factor $\kappa$.

Importantly, the geometric projection is performed only when the original
input vector violates the target constraint,
\[
H_0=H(\vy)>l.
\]
When
\[
H_0\leq l,
\]
the input is already feasible and no geometric projection toward the
boundary is required.
\subsection{Common Active-Set Threshold for PLCP and PECP}
\label{sec:common_threshold}

Before introducing the one-sided PLCP and PECP algorithms, we derive the
common threshold used to identify their active support.

The key point is that the support identification is determined solely by
the CAI boundary condition and is therefore independent of the subsequent
geometric projection. Consequently, PLCP and PECP share exactly the same
active-set threshold.

\subsubsection{CAI boundary on the active support}
\label{active-set}

Let
\[
\vx\in\mathbb R_+^n
\]
and assume that the current active support contains $\nu$ coefficients.
Restricting $\vx$ to this support, the CAI boundary is defined by
\[
H(\vx)=l,
\]
where
\[
H(\vx)
=
\frac{\|\vx\|_1^2}{\|\vx\|_2^2}.
\]

Let
\[
S_1=\|\vx\|_1,
\qquad
S_2=\|\vx\|_2^2,
\qquad
H=\frac{S_1^2}{S_2}.
\]

Consider the centered vector
\[
\vz
=
\vx-\alpha\mathbf 1_\nu,
\]
where $\alpha$ denotes the threshold associated with the current active
support.

Its first two moments are
\[
\|\vz\|_1
=
S_1-\nu\alpha,
\]
and
\[
\|\vz\|_2^2
=
S_2-2\alpha S_1+\nu\alpha^2.
\]

Imposing the CAI boundary condition
\[
H(\vz)=l
\]
gives
\[
\frac{(S_1-\nu\alpha)^2}
{S_2-2\alpha S_1+\nu\alpha^2}
=
l.
\]

Therefore,
\[
\nu(\nu-l)\alpha^2
-
2(\nu-l)S_1\alpha
+
S_1^2-lS_2
=
0.
\]

Using
\[
S_2=\frac{S_1^2}{H},
\]
we obtain
\[
\nu(\nu-l)\alpha^2
-
2(\nu-l)S_1\alpha
+
S_1^2
\left(
1-\frac{l}{H}
\right)
=
0.
\]

Dividing by $\nu(\nu-l)$ gives
\[
\alpha^2
-
2\frac{S_1}{\nu}\alpha
+
\frac{S_1^2}{\nu(\nu-l)}
\left(
1-\frac{l}{H}
\right)
=
0.
\]

Solving this quadratic equation gives
\[
\alpha
=
\frac{S_1}{\nu}
\left[
1
\pm
\sqrt{
1-
\frac{\nu}{\nu-l}
\left(
1-\frac{l}{H}
\right)
}
\right].
\]

The term inside the square root simplifies as
\[
1-
\frac{\nu}{\nu-l}
\left(
1-\frac{l}{H}
\right)
=
\frac{l(\nu-H)}
{H(\nu-l)}.
\]

Hence,
\[
\alpha
=
\frac{S_1}{\nu}
\left(
1
\pm
\sqrt{
\frac{l(\nu-H)}
{H(\nu-l)}
}
\right).
\]

The relevant root for the active-set selection is the smaller one,
leading to the common threshold
\[
\boxed{
\alpha
=
\frac{\|\vx\|_1}{\nu}
\left(
1-
\sqrt{
\frac{
l\bigl(\nu-H(\vx)\bigr)
}{
H(\vx)(\nu-l)
}
}
\right).
}
\]

This threshold determines which coefficients remain active:
\[
\boxed{
x_i
\gets
\begin{cases}
x_i, & x_i\geq\alpha,\\
0,   & x_i<\alpha.
\end{cases}
}
\]

The active-set procedure is repeated until the support size $\nu$
stabilizes.

\subsubsection{Why the threshold is common to PLCP and PECP}

The derivation of $\alpha$ uses only the CAI boundary equation
\[
H=l
\]
and the current active support size $\nu$.

It does not involve either the PLCP extrapolation coefficient $\lambda$
or the PECP Euclidean projection parameter $t$.

Therefore, the threshold is independent of the particular geometric path
used to reach the CAI boundary:
\[
\boxed{
\alpha_{\mathrm{PLCP}}
=
\alpha_{\mathrm{PECP}}
=
\alpha.
}
\]

This provides the theoretical justification for using a single
active-set selection procedure for both methods.

Once the support has stabilized, the two methods differ only in their
geometric projection:
\[
\boxed{
\begin{array}{c}
\text{CAI boundary condition}
\\[1mm]
H=l
\\
\Downarrow
\\
\text{common threshold }\alpha
\\
\Downarrow
\\
\text{common active support }\mathcal A
\\
\Downarrow
\\[1mm]
\begin{cases}
\text{PLCP: affine projection using }\lambda,\\[1mm]
\text{PECP: Euclidean projection using }t.
\end{cases}
\end{array}
}
\]

\subsection{Derivation of the PLCP Extrapolation Factor}
\label{PLCP_lambda}

Once the active support has stabilized, let
\[
A=\{i:x_i>0\},
\qquad
|A|=\nu,
\]
and let $\alpha$ denote the final threshold obtained from the
CAI boundary condition.

On the active support, define the diagonal reference vector
\[
\mathbf d
=
\frac{\|\mathbf x\|_1}{\nu}\mathbf 1_A.
\]

The shifted vector associated with the final threshold is
\[
\widetilde{\mathbf x}
=
\mathbf x-\alpha\mathbf 1_A.
\]
By construction of the threshold $\alpha$,
\[
H(\widetilde{\mathbf x})=l.
\]

Since
\[
\mathbf 1_A
=
\frac{\nu}{\|\mathbf x\|_1}\mathbf d,
\]
the shifted vector can be rewritten as
\[
\widetilde{\mathbf x}
=
\mathbf x
-
\frac{\alpha\nu}{\|\mathbf x\|_1}\mathbf d.
\]

Let
\[
c
=
\frac{\alpha\nu}{\|\mathbf x\|_1}.
\]
Then
\[
\widetilde{\mathbf x}
=
\mathbf x-c\mathbf d.
\]

The PLCP solution is defined by the affine extrapolation
\[
\widehat{\mathbf x}_{\mathrm{PLCP}}
=
\lambda\mathbf x+(1-\lambda)\mathbf d.
\]

Choosing
\[
\lambda=\frac{1}{1-c}
\]
gives
\[
1-\lambda
=
-\frac{c}{1-c},
\]
and therefore
\[
\widehat{\mathbf x}_{\mathrm{PLCP}}
=
\frac{1}{1-c}\mathbf x
-
\frac{c}{1-c}\mathbf d
=
\frac{\mathbf x-c\mathbf d}{1-c}.
\]

Using
\[
\mathbf x-c\mathbf d
=
\widetilde{\mathbf x},
\]
we obtain
\[
\widehat{\mathbf x}_{\mathrm{PLCP}}
=
\lambda\widetilde{\mathbf x}.
\]

Since the CAI score is scale-invariant,
\[
H(\widehat{\mathbf x}_{\mathrm{PLCP}})
=
H(\widetilde{\mathbf x})
=
l.
\]
Hence, the affine extrapolation lies on the CAI boundary.

Finally,
\[
\boxed{
\lambda
=
\frac{1}{
1-\dfrac{\alpha\nu}{\|\mathbf x\|_1}
}.
}
\]

Substituting the closed-form threshold
\[
\alpha
=
\frac{\|\mathbf x\|_1}{\nu}
\left(
1-
\sqrt{
\frac{
l\bigl(\nu-H(\mathbf x)\bigr)
}{
H(\mathbf x)(\nu-l)
}
}
\right)
\]
yields
\[
\boxed{
\lambda
=
\sqrt{
\frac{
H(\mathbf x)(\nu-l)
}{
l\bigl(\nu-H(\mathbf x)\bigr)
}
}.
}
\]

Thus, once the final active support and the threshold $\alpha$
have been identified, the PLCP extrapolation factor is obtained
directly in closed form, without any additional optimization.

\subsubsection{PECP from Euclidean-Norm Minimization}
\label{PECPfrob}

Let $\mathbf y$ denote the input vector restricted to the final
active support. The PECP solution is defined by
\[
\widehat{\mathbf x}_{\rm PECP}
=
\arg\min_{\mathbf x}
\frac12\|\mathbf x-\mathbf y\|_2^2
\quad
\text{subject to } H(\mathbf x)=l.
\]

\paragraph{Axial-orthogonal decomposition.}

Let $\mathcal A$ denote the final active support and let
\[
|\mathcal A|=\nu.
\]

In the corresponding active subspace, define the normalized diagonal
axis as
\[
\boxed{
\vu
=
\frac{1}{\sqrt{\nu}}\mathbf{1}_{\mathcal A}.
}
\]

The input vector restricted to the active support can be decomposed as
\[
\boxed{
\vy_{\mathcal A}
=
s\vu+r\vv,
}
\]
where
\[
\boxed{
s
=
\vy_{\mathcal A}^{\top}\vu
=
\frac{\|\vy_{\mathcal A}\|_1}{\sqrt{\nu}},
}
\]
and
\[
\boxed{
r
=
\left\|
\vy_{\mathcal A}-s\vu
\right\|_2.
}
\]

For $r>0$, the unit orthogonal direction is
\[
\boxed{
\vv
=
\frac{
\vy_{\mathcal A}-s\vu
}{
r
}.
}
\]

By construction,
\[
\|\vu\|_2=\|\vv\|_2=1,
\qquad
\vu^\top\vv=0.
\]

\paragraph{CAI boundary.}

The CAI boundary in the active subspace is defined by
\[
\boxed{
H(\vx)=l.
}
\]

Let $\theta_l$ denote the half-opening angle of the CAI cone in this
active subspace. Since
\[
\cos\theta_l
=
\sqrt{\frac{l}{\nu}},
\]
we obtain
\[
\tan^2\theta_l
=
\frac{1-\cos^2\theta_l}{\cos^2\theta_l}
=
\frac{\nu-l}{l}.
\]

Hence,
\[
\boxed{
q
=
\tan\theta_l
=
\sqrt{\frac{\nu-l}{l}}.
}
\]

Any point on the CAI boundary generator contained in the plane spanned
by $\vu$ and $\vv$ can therefore be parameterized as
\[
\boxed{
\vx(t)
=
t\vu+tq\vv
=
t(\vu+q\vv).
}
\]

Indeed, its axial component is $t$, whereas the norm of its orthogonal
component is $tq$, and therefore
\[
\frac{
\|\vx(t)-(\vx(t)^\top\vu)\vu\|_2
}{
\vx(t)^\top\vu
}
=
q.
\]

\paragraph{Euclidean-norm minimization.}
PECP selects the point on this CAI boundary generator that minimizes
the Euclidean distance to the input vector.

\[
\boxed{
t^\star
=
\arg\min_{t\geq0}
\frac{1}{2}
\left\|
\vy_{\mathcal A}
-
t(\vu+q\vv)
\right\|_2^2.
}
\]

Using the decomposition
\[
\vy_{\mathcal A}
=
s\vu+r\vv,
\]
we obtain
\[
\vy_{\mathcal A}
-
t(\vu+q\vv)
=
(s-t)\vu
+
(r-qt)\vv.
\]

Since $\vu$ and $\vv$ are orthonormal,
\[
\left\|
\vy_{\mathcal A}
-
t(\vu+q\vv)
\right\|_2^2
=
(s-t)^2+(r-qt)^2.
\]

Thus, the PECP problem reduces to the one-dimensional minimization
\[
\boxed{
t^\star
=
\arg\min_{t\geq0}
\frac{1}{2}
\left[
(s-t)^2
+
(r-qt)^2
\right].
}
\]

Define
\[
\phi(t)
=
\frac{1}{2}
\left[
(s-t)^2
+
(r-qt)^2
\right].
\]

Differentiating with respect to $t$ gives
\[
\phi'(t)
=
-(s-t)
-
q(r-qt).
\]

The optimality condition
\[
\phi'(t)=0
\]
therefore yields
\[
-(s-t)-q(r-qt)=0.
\]

Rearranging,
\[
s+qr
=
(1+q^2)t.
\]

Consequently,
\[
\boxed{
t^\star
=
\frac{s+qr}{1+q^2}.
}
\]

Moreover,
\[
\phi''(t)
=
1+q^2>0,
\]
so that $t^\star$ is the unique minimizer.

Hence, the PECP solution is the Euclidean projection onto the
corresponding CAI boundary generator. In the $(s,r)$ plane, this
reduces to the orthogonal projection of the point $(s,r)$ onto the
boundary line $r=qs$.

\paragraph{Closed-form PECP solution.}

The PECP solution is therefore
\[
\boxed{
\widehat{\vx}_{\mathrm{PECP}}
=
t^\star(\vu+q\vv),
}
\]
with
\[
\boxed{
t^\star
=
\frac{s+qr}{1+q^2}.
}
\]

Equivalently,
\[
\boxed{
\widehat{\vx}_{\mathrm{PECP}}
=
\frac{s+qr}{1+q^2}
(\vu+q\vv).
}
\]

Returning to the matrix representation gives
\[
\boxed{
\widehat{\mathbf{X}}_{\mathrm{PECP}}
=
\operatorname{unvec}
\left[
\frac{s+qr}{1+q^2}
(\vu+q\vv)
\right].
}
\]

Hence, the PECP solution is the Euclidean projection onto the
corresponding CAI boundary generator. In the $(s,r)$ plane,
this reduces to the orthogonal projection of the point $(s,r)$
onto the boundary line $r=qs$.

\[
\boxed{
r=qs.
}
\]

\paragraph{Relation between PLCP and PECP}

Let
\[
\vu
=
\frac{1}{\sqrt{\nu}}\mathbf 1_{\mathcal A}
\]
denote the normalized diagonal axis of the active subspace, and let
$\vv$ denote the corresponding unit orthogonal direction.

The CAI boundary is characterized by
\[
q
=
\tan\theta_l
=
\sqrt{\frac{\nu-l}{l}}.
\]

Therefore, any point lying on the corresponding CAI boundary generator
can be written as
\[
\rho(\vu+q\vv),
\qquad \rho\geq0.
\]

The PLCP solution lies on this generator and can therefore be written as
\[
\boxed{
\widehat{\vx}_{\mathrm{PLCP}}
=
s_{\mathrm{PLCP}}
(\vu+q\vv),
}
\]
where $s_{\mathrm{PLCP}}$ denotes the axial coordinate of the PLCP
solution.

Similarly, from the Euclidean projection derived for PECP,
\[
\boxed{
\widehat{\vx}_{\mathrm{PECP}}
=
t(\vu+q\vv),
}
\]
where
\[
t
=
\frac{s+qr}{1+q^2}.
\]

Here, $s$ and $r$ denote respectively the axial and orthogonal
coordinates of the vector to be projected:
\[
s
=
\vy_{\mathcal A}^{\top}\vu,
\]
and
\[
r
=
\left\|
\vy_{\mathcal A}-s\vu
\right\|_2.
\]

Since both solutions lie on the same CAI boundary generator, they are
collinear. Indeed,
\[
\widehat{\vx}_{\mathrm{PECP}}
=
t(\vu+q\vv)
\]
and
\[
\widehat{\vx}_{\mathrm{PLCP}}
=
s_{\mathrm{PLCP}}(\vu+q\vv).
\]

Consequently,
\[
\widehat{\vx}_{\mathrm{PECP}}
=
\frac{t}{s_{\mathrm{PLCP}}}
\widehat{\vx}_{\mathrm{PLCP}}.
\]

Therefore,
\begin{equation}
\label{kappa}
\boxed{
\kappa
=
\frac{t}{s_{\mathrm{PLCP}}}.
}
\end{equation}

Hence, the PECP and PLCP solutions satisfy
\[
\boxed{
\widehat{\vx}_{\mathrm{PECP}}
=
\kappa\,
\widehat{\vx}_{\mathrm{PLCP}}.
}
\]
Thus, the two methods differ only by a radial scaling once the common
active support has been identified.

This yields the unified interpretation
\[
\boxed{
\begin{array}{ccc}
& \text{Common active-set selection} & \\[1mm]
& \alpha,\;\nu & \\[1mm]
& \Downarrow & \\[2mm]
H_0>l
&
\begin{cases}
\text{PLCP: affine projection using }\lambda,\\
\text{PECP: Euclidean projection using }t\text{ or }\kappa,
\end{cases}
\\[5mm]
H_0\leq l
&
\text{no geometric projection}.
\end{array}
}
\]

Therefore, PLCP and PECP select exactly the same conic
direction. They differ only by a radial scaling along that cone generator:
PLCP reaches the boundary through the affine path joining $\vy$ to the
diagonal point $\vd$, whereas the Euclidean projection reaches the same
generator through the shortest-distance path.

\subsection{Euclidean characterization of the bilevel PECP}
\label{PECP-Frob}
\begin{proposition}[Euclidean characterization of the bilevel PECP]
Let
\[
Y=[y_1,\ldots,y_n]\in\mathbb{R}^{m\times n},
\qquad
v=(\|y_1\|_2,\ldots,\|y_n\|_2),
\]
and let
\[
\widehat u=P_l^{\mathrm{PECP}}(v)
\]
denote the Euclidean CAI projection of the vector of column norms.
Define
\[
x_j=P_{\widehat u_j}^{2}(y_j),
\qquad j=1,\ldots,n.
\]
Then, provided $0\leq\widehat u_j\leq\|y_j\|_2$, the resulting
matrix $X=[x_1,\ldots,x_n]$ minimizes the Frobenius distance to $Y$
among matrices whose vector of column $\ell_2$ norms satisfies the
corresponding CAI constraint.
\end{proposition}

\begin{proof}
For any prescribed vector of column radii
$u=(u_1,\ldots,u_n)$, the minimization of the Frobenius distance
separates columnwise:
\[
\min_{\|x_j\|_2\leq u_j}
\|Y-X\|_F^2
=
\sum_{j=1}^n
\min_{\|x_j\|_2\leq u_j}
\|y_j-x_j\|_2^2.
\]
The Euclidean projection onto the $\ell_2$ ball gives
\[
x_j=P_{u_j}^{2}(y_j).
\]
Since $0\leq u_j\leq v_j=\|y_j\|_2$,
\[
\|y_j-x_j\|_2=v_j-u_j.
\]
Therefore,
\[
\|Y-X\|_F^2
=
\sum_{j=1}^n(v_j-u_j)^2
=
\|v-u\|_2^2.
\]
Consequently, minimizing the Frobenius distance in matrix space is
equivalent to minimizing the Euclidean distance between the vectors
of column norms:
\[
\min_X\frac12\|Y-X\|_F^2
=
\min_{u:\,H(u)\leq l}
\frac12\|v-u\|_2^2.
\]
Since $\widehat u=P_l^{\mathrm{PECP}}(v)$ is the Euclidean CAI
projection of $v$, the matrix obtained from
$x_j=P_{\widehat u_j}^{2}(y_j)$ is the corresponding exact
Frobenius projection.
\end{proof}

\subsection{Laguerre's style and Samuelson's style bound }
\label{zerosum-proof}
\noindent

Let $y_1,\dots,y_n$ be positive real numbers, and set
\[
 l_1:=\sum_{i=1}^n y_i,
 \qquad
 l_2:=\sqrt{\sum_{i=1}^n y_i^2},
 \qquad
 H:=\left(\frac{l_1}{l_2}\right)^2.
\]
Fix a positive number $l<n$, and define
\begin{equation}
    \label{eq:musigma}
\mu:=\frac{l_1}{n},
\qquad
\sigma^2:=\frac1n\sum_{i=1}^n (y_i-\mu)^2.
\end{equation}
We will prove that the number of indices $i$ such that
\[
 y_i\le\frac{l_1}{n}\left(1-\sqrt{\frac{l(n-H)}{H(n-l)}}\right)
\]
is at most $n-l$.

First, remark that, according to the following lemma, this threshold can be rewritten in terms of the \emph{standard score} (aka \emph{z-score}):
\begin{equation}
\label{eq:zscore}
\frac{y_i - \mu}{\sigma} \le - \sqrt{\frac{l}{n - l}}.
\end{equation}
\begin{lemma}
    The numbers $\mu$ and $\sigma$ defined in equation~\ref{eq:musigma} satisfy
\begin{equation}
    \label{eq:musigmaH}
\sigma^2
=\mu^2\!\left(\frac{n-H}{H}\right).
\end{equation}

\begin{proof}
We have
\begin{align*}
\sigma^2
 &= \frac1n\sum_{i=1}^n y_i^2-\mu^2 = \frac{l_2^2}{n}-\mu^2.
\end{align*}
Using $H=(l_1/l_2)^2$ gives $l_2^2=l_1^2/H=n^2\mu^2/H$, which gives equation~\ref{eq:musigmaH}.
\end{proof}
\end{lemma}

\begin{lemma}
\label{lemma:general_inequality}
Let
\[
r :=\#\left\{ i:\; y_i \le \frac{l_1}{n} \left( 1-\sqrt{\frac{l(n-H)}{H(n-l)}} \right) \right\}
  = \#\left\{ i:\; y_i\le \mu-t \right\},
\]
where
\[
 t:=\mu\sqrt{\frac{l(n-H)}{H(n-l)}}.
\]
Then
\[
r\le \frac{n\sigma^2}{\sigma^2+t^2}.
\]
\end{lemma}

As detailed in \cite{Jensen1999}, there are two approaches to prove such Laguerre-Samuelson style inequality.
A first one, following Laguerre's style uses Cauchy-Schwarz and a second one, following Samuelson's style uses a clustering argument.
We provide both of them below.

\begin{proof}[Proof in Laguerre style]
Write the deviations from the mean as
\[
 d_i:=y_i-\mu,
\qquad i=1,\dots,n.
\]
Then
\[
\sum_{i=1}^n d_i=0 \text{ (zero sum)},
\qquad
\sum_{i=1}^n d_i^2=n\sigma^2.
\]

We derive the desired result by bounding the total variation around
the mean.
Among the numbers $d_i$, exactly $r$ of them satisfy $d_i\le -t$.
Let
\[
A:=\{i:\ d_i\le -t\},
\qquad |A|=r.
\]
Set
\begin{equation}
\label{eq:S}
    S_A:=\sum_{i\in A} d_i,
\qquad
S_{\overline{A}}:=\sum_{i\notin A} d_i.
\end{equation}
Since the total sum is zero, we have $S_{\overline{A}}=-S_A$.
Moreover, because each $i\in A$ satisfies $d_i\le -t$, we get
\[
S_A\le -rt,
\qquad\text{hence}\qquad
S_{\overline{A}}\ge rt.
\]
By Cauchy--Schwarz applied to the set $\overline{A}$,
\[
S_{\overline{A}}^2
\le |\overline{A}|\sum_{i\notin A} d_i^2
=(n-r)\sum_{i\notin A} d_i^2.
\]
Therefore
\[
\sum_{i\notin A} d_i^2\ge \frac{S_{\overline{A}}^2}{n-r}\ge \frac{r^2t^2}{n-r}.
\]
Also, for $i\in A$ we have $d_i^2\ge t^2$, so
\[
\sum_{i\in A} d_i^2\ge rt^2.
\]
Adding the two estimates gives
\begin{align}
\notag
n\sigma^2
 &= \sum_{i=1}^n d_i^2
  = \sum_{i\in A} d_i^2+\sum_{i\notin A} d_i^2 \\
  \label{eq:sigmarnt}
 &\ge rt^2+\frac{r^2t^2}{n-r}
  = \frac{rnt^2}{n-r}.
\end{align}
Rearranging,
\[
r\le \frac{n\sigma^2}{\sigma^2+t^2}.
\]
\end{proof}

\begin{proof}[Proof in Samuelson style]
This proof uses the remark ``putting 2 measurements at their common mean reduces the standard deviation while leaving the mean invariant'' from Samuelson's paper~\cite{samuelson1968deviant}.
Consider the clustering operation replacing all samples in $A$ by their mean $\mu_A$ and all samples
in $\overline{A}$ by their mean $\mu_{\overline{A}}$; so let
$y'_i = \mu_A$ if $i \in A$ and $y'_i = \mu_{\overline{A}}$ otherwise.
Let $\mu'$ and $\sigma'$ be their mean and standard deviation computed as in
Equation~\ref{eq:musigma}. Define $S'$ as in Equation \ref{eq:S}.
Using Samuelson's remark, we know that $\sigma' \le \sigma$.
We have $S'_A + S'_{\overline{A}} = 0$ so $S'_{\overline{A}} = -S'_A$.
As $d_i < -t$ for all $i \in A$, we have $S'_A \le -rt$. Therefore
\[
(\sigma')^2 = \frac{1}{n}(S_A^2/r + S_{\overline{A}}^2/(n-r)) = \frac{S_A^2}{n} (1/r + 1/(n-r)) = \frac{S_A^2}{r(n-r)} \ge \frac{rt^2}{n-r}.
\]
This is Equation~\ref{eq:sigmarnt} so the last step is shared with the Laguerre style proof.
\end{proof}

\begin{theorem}
$r\le n-l$.
\end{theorem}

\begin{proof}
By equation~\ref{eq:musigmaH},
\begin{align*}
\frac{t^2}{\sigma^2}
 &= \frac{\mu^2\,\dfrac{l(n-H)}{H(n-l)}}{\mu^2\,\dfrac{n-H}{H}} = \frac{l}{n-l}.
\end{align*}
Lemma \ref{lemma:general_inequality} gives us
\[
r\le \frac{n\sigma^2}{\sigma^2+t^2}.
\]
Substituting $t^2/\sigma^2=l/(n-l)$ yields
\begin{align*}
r
 &\le \frac{n}{1+t^2/\sigma^2} \\
 &= \frac{n}{1+\dfrac{l}{n-l}} \\[6pt]
 &= \frac{n(n-l)}{n} \\
 &= n-l.
\end{align*}
\end{proof}

\subsection{Euclidean characterization of the bilevel PECP}
\label{ƒrobenius}
\begin{proposition}[Euclidean characterization of the bilevel PECP]
Let
\[
Y=[y_1,\ldots,y_n]\in\mathbb{R}^{m\times n},
\qquad
v=(\|y_1\|_2,\ldots,\|y_n\|_2),
\]
and let
\[
\widehat u=P_l^{\mathrm{PECP}}(v)
\]
denote the Euclidean CAI projection of the vector of column norms.
Define
\[
x_j=P_{\widehat u_j}^{2}(y_j),
\qquad j=1,\ldots,n.
\]
Then, provided $0\leq\widehat u_j\leq\|y_j\|_2$, the resulting
matrix $X=[x_1,\ldots,x_n]$ minimizes the Frobenius distance to $Y$
among matrices whose vector of column $\ell_2$ norms satisfies the
corresponding CAI constraint.
\end{proposition}

\begin{proof}
For any prescribed vector of column radii
$u=(u_1,\ldots,u_n)$, the minimization of the Frobenius distance
separates columnwise:
\[
\min_{\|x_j\|_2\leq u_j}
\|Y-X\|_F^2
=
\sum_{j=1}^n
\min_{\|x_j\|_2\leq u_j}
\|y_j-x_j\|_2^2.
\]
The Euclidean projection onto the $\ell_2$ ball gives
\[
x_j=P_{u_j}^{2}(y_j).
\]
Since $0\leq u_j\leq v_j=\|y_j\|_2$,
\[
\|y_j-x_j\|_2=v_j-u_j.
\]
Therefore,
\[
\|Y-X\|_F^2
=
\sum_{j=1}^n(v_j-u_j)^2
=
\|v-u\|_2^2.
\]
Consequently, minimizing the Frobenius distance in matrix space is
equivalent to minimizing the Euclidean distance between the vectors
of column norms:
\[
\min_X\frac12\|Y-X\|_F^2
=
\min_{u:\,H(u)\leq l}
\frac12\|v-u\|_2^2.
\]
Since $\widehat u=P_l^{\mathrm{PECP}}(v)$ is the Euclidean CAI
projection of $v$, the matrix obtained from
$x_j=P_{\widehat u_j}^{2}(y_j)$ is the corresponding exact
Frobenius projection.
\end{proof}

\subsection{Comparison of thresholding iteration with literature}
\label{an:lit}

\subsubsection{Comparison with Torrent family}
\label{sec:bhatia}

Hard-thresholding iterations of the form Equation 
\ref{hard_threshold} are also at the heart of the T\textsc{orrent} family of algorithms introduced by \cite{bhatia2015} for Robust Least Squares Regression (RLSR). It is therefore instructive to delineate precisely how our PLCP iteration relates to, and differs from, that line of work.

\paragraph{Problem setting.}
T\textsc{orrent} addresses the regression model $\vy = \mX^\top \vw^\star + \vb$, where $\vb \in \mathbb{R}^n$ is a sparse but adversarially chosen corruption vector with $\|\vb\|_0 \le \alpha n$. The goal is to recover $\vw^\star$ exactly despite an unknown set of corrupted responses. The algorithm alternates between (i) a hard-thresholding step that selects an active set $S_t = \mathrm{HT}(\vr^t,(1-\beta)n)$ of points with the smallest residuals, and (ii) a fit step (least squares for T\textsc{orrent}-FC, a gradient step for T\textsc{orrent}-GD, or a hybrid policy for T\textsc{orrent}-HYB) that re-estimates $\vw$ on $S_t$.

By contrast, the PLCP iteration solves a Euclidean projection onto a structured sparsity cone arising from neural network compression. There is no statistical model, no corruption, and no regressor to recover: the iteration acts directly on the entries of $\vx$ (or on the vector of column $\ell_2$-norms, in the bilevel variant) and the active support is determined by a magnitude threshold $\alpha(\vx)$ that is itself a function of the current iterate.

\paragraph{Form of the thresholding step.}
The hard-thresholding operator used in \cite{bhatia2015} selects a fixed-cardinality set of indices (the $(1-\beta)n$ smallest residuals), so the active-set size is held constant across iterations. In our setting, the threshold $\alpha(\vx^{(t)})$ is computed from the current iterate via the cone characterization, and the cardinality of the active support $|\{i : x_i^{(t)} \ge \alpha(\vx^{(t)})\}|$ is itself updated at each step. The PLCP iteration is therefore an \emph{adaptive-cardinality} thresholding scheme, whereas T\textsc{orrent} is a \emph{fixed-cardinality} one.

\paragraph{Convergence guarantee.}
The two analyses also differ in nature. T\textsc{orrent} establishes a geometric (linear) convergence rate $\|\vw^t - \vw^\star\|_2 \le C \rho^t$ under deterministic subset strong convexity/smoothness (SSC/SSS) assumptions on $\mX$, which hold with high probability for sub-Gaussian designs; the iteration count to reach an $\varepsilon$-accurate solution scales as $\mathcal{O}(\log(\|\vb\|_2/(\sqrt n\,\varepsilon)))$, and exact recovery is only reached in the limit. The PLCP iteration, on the other hand, enjoys a \emph{finite-time} guarantee: the active support is monotonically non-increasing along iterations, hence the loop reaches a fixed point in at most $n$ steps, with no probabilistic or design assumption on the input vector. Empirically (see Figure~\ref{active}), in most cases the active-set size collapses exponentially fast toward its limit, well before the worst-case $n$-iteration bound.

\subsubsection{Comparison with the Luo--Tseng convergence framework}
\label{sec:luo1992}

A second, more classical line of work on the convergence of active-set iterations is the analysis of \cite{luo1992} for coordinate descent on problems of the form
\[
\min_{\vx \in \mathcal{X}} \; g(\mE \vx) + \langle \vb, \vx \rangle,
\]
where $\mathcal{X}$ is a (possibly unbounded) box, $\mE$ has no zero column, and $g$ is strictly convex and twice differentiable on its effective domain. Their main result is that the iterates generated by coordinate descent converge to an optimal solution at a \emph{linear} rate, without requiring that the cost function have bounded level sets or be strictly convex on the whole space. As a corollary, they obtain the first linear-convergence guarantee for the dual iterates of several methods for symmetric monotone linear complementarity problems, matrix balancing, and entropy optimization.

\paragraph{Common ground.}
The PLCP iteration shares with the Luo--Tseng setting two features that are central to their analysis. First, the constraint set involved in our projection is a polyhedral cone -- in particular a Cartesian product of half-lines after the support has been selected -- so the underlying feasibility set has the same box-like geometry as $\mathcal{X}$. Second, the optimality conditions of the conic projection can be written as a linear complementarity system (positivity, primal feasibility, and complementary slackness on the active support), which is precisely the class of problems for which \citet{luo1992} establish linear convergence of coordinate descent. In that sense, the PLCP optimality conditions live in the LCP universe that Luo--Tseng's analysis targets.

\paragraph{Where the analyses differ.}
The differences are equally informative. \cite{luo1992} update \emph{one coordinate at a time} (in an almost-cyclic or Gauss--Southwell order) and rely on an error-bound inequality to obtain a geometric rate of convergence on the iterates. The PLCP iteration, in contrast, performs a \emph{global} active-support update at each step: all coordinates below the current threshold $\alpha(\vx^{(t)})$ are zeroed simultaneously, and the threshold itself is recomputed in closed form from the resulting support. Because the support is a discrete object and only shrinks across iterations, this block update yields a strictly stronger guarantee than the linear rate of \cite{luo1992}: the iteration reaches its fixed point in at most $n$ steps, with no error-bound or strict-convexity assumption. Empirically, the active-support size further decays exponentially fast (Figure~\ref{active}), so that the worst-case $n$-iteration bound is almost never tight.

\subsubsection{Comparison with Least Median of Squares regression}
\label{sec:rousseeuw1984}

Predating the modern hard-thresholding literature by several decades, \cite{rousseeuw1984} introduced \emph{Least Median of Squares} (LMS) regression as a high-breakdown-point alternative to ordinary least squares:
\[
\hat{\vw}_{\mathrm{LMS}} \;\in\; \arg\min_{\vw \in \mathbb{R}^p} \; \mathrm{med}_{i=1,\dots,n}\bigl( y_i - \vx_i^\top \vw \bigr)^2.
\]
Because the median is, up to ties, the $\lfloor n/2\rfloor$-th order statistic of the squared residuals, LMS implicitly identifies the half of the observations on which the regressor fits best and ignores the rest. In this sense it is the conceptual ancestor of T\textsc{orrent}'s hard-thresholding step and, more broadly, of every active-set scheme that discards the points with the largest residuals.

\paragraph{Selection mechanism.}
LMS and PLCP share the same underlying primitive -- ranking quantities (residuals, respectively coefficient magnitudes) and keeping only those above a data-adaptive threshold -- but they differ in three ways:
\emph{(i)} the threshold in LMS is fixed at the median rank ($50\%$ of the data), giving the estimator its maximal breakdown point $\varepsilon^\star = 1/2$; in PLCP the threshold $\alpha(\vx)$ is computed in closed form from the cone-alignment constraint and the active-set size is therefore data-driven rather than prescribed;
\emph{(ii)} LMS ranks \emph{squared residuals} of a regression fit, whereas PLCP ranks the entries of the iterate itself (or the column $\ell_2$-norms in the bilevel variant);
\emph{(iii)} LMS only specifies the optimization \emph{objective}; PLCP, like T\textsc{orrent}, additionally specifies an \emph{iterative procedure} for finding the active support, and we prove that this procedure terminates in finitely many steps.

\paragraph{Robustness vs.\ deterministic termination.}
The theoretical guarantee provided by \citet{rousseeuw1984} is statistical and worst-case: LMS attains the highest possible breakdown point among regression-equivariant estimators, meaning that up to almost half of the observations can be corrupted arbitrarily without the estimator drifting to infinity. Our guarantee for PLCP, by contrast, is algorithmic: there is no notion of breakdown point because the input is a deterministic vector to be projected, not a noisy sample; instead, the iteration is shown to reach a fixed point in at most $n$ steps for any input. The two analyses therefore answer different questions about the same family of thresholding ideas -- "how much corruption can the answer survive?" versus "how fast does the iteration stop?" -- and are best read as complementary rather than competing.

\paragraph{Computational profile.}
A practical contrast worth noting is that exact LMS is NP-hard in $p$ and is typically solved by random subsampling heuristics with no global guarantee, whereas the PLCP iteration is a deterministic procedure with closed-form updates and worst-case complexity $\mathcal{O}(Kn)$, $K \le n$. Combined with the empirical observation that $K$ remains small (Figure~\ref{active}), this makes PLCP attractive as the structured-sparsity analogue of the thresholding-by-ranking idea that LMS first crystallized in the robust-regression setting.

\subsubsection{Comparison with Huber's M-estimators and IRLS }
\label{sec:huber1981}

The most foundational reference for the convergence of "trim and refit" iterations is the theory of M-estimators developed by \cite{huber1981} in Chapter 3 of \emph{Robust Statistics}. An M-estimator is defined as the solution of
\[
\hat{\vw}_{\mathrm{M}} \;\in\; \arg\min_{\vw} \;\sum_{i=1}^n \rho\bigl(y_i - \vx_i^\top \vw\bigr),
\]
for a loss $\rho$ chosen to attenuate the influence of large residuals, and it is computed in practice by an Iteratively Reweighted Least Squares (IRLS) scheme of the form
\[
\vw^{(t+1)} \;=\; \arg\min_{\vw} \sum_{i=1}^n w_i\bigl(\vw^{(t)}\bigr)\,\bigl(y_i - \vx_i^\top \vw\bigr)^2,
\qquad w_i(\vw) \;=\; \frac{\psi(r_i(\vw))}{r_i(\vw)},
\]
where $\psi=\rho'$ is the influence function. Huber gives the foundational analysis of when this iteration converges and to what.

\paragraph{PLCP as IRLS with binary weights.}
The PLCP iteration fits squarely inside this framework once one notices that the hard-thresholding update Equation \ref{hard_threshold} is precisely IRLS with a \emph{binary} weight function
\[
w_i\bigl(\vx^{(t)}\bigr) \;=\; \mathbf{1}\!\left[ x_i^{(t)} \ge \alpha\bigl(\vx^{(t)}\bigr) \right] \in \{0,1\},
\]
corresponding to the discontinuous "hard" influence function $\psi(u) = u\,\mathbf{1}[u\ge\alpha]$ rather than Huber's clipped-linear $\psi$. The pattern "compute the (cone-adapted) statistic, threshold, recompute" is, in Huber's terminology, a degenerate redescending M-iteration in which observations either count fully or not at all. Standard IRLS with continuous weights interpolates between these extremes; PLCP and T\textsc{orrent} both correspond to the binary-weight limit.

\paragraph{What Huber's analysis buys us, and where it stops.}
Huber's convergence proofs for IRLS rely on smoothness of $\rho$ and on the resulting majorization-minimization structure of the iteration. They give monotone descent of an explicit objective and asymptotic convergence under mild conditions. Specializing them to a discontinuous binary $\psi$ formally falls outside their regularity assumptions, but the underlying intuition -- that the active-set update is a majorization step on a piecewise-quadratic surrogate -- still applies and explains why such iterations tend to converge in practice. Our analysis can be read as supplying the missing piece in this degenerate regime: rather than invoking smoothness, we use the combinatorial fact that the active support is a monotone subset sequence to conclude finite-time termination in at most $n$ steps. In this sense, the PLCP convergence result is the natural counterpart, on the structured-sparsity side, of Huber's IRLS convergence theory: same iterative template, different mathematical tool for closing the proof.

\subsection{Basic Inference analysis}
\label{inference}

We provide a basic analysis of the computational cost of using PLCP
within a BERT-like attention block \cite{BERT_I}. \\

In this simple illustrative setting, token importance is approximated
by the embedding norm, and tokens associated with smaller norms are
preferentially removed.
Let the original sequence length be $n$ and the head dimension be $d$. 
When the quadratic attention operations dominate the computational
cost, the leading-order complexity becomes

\[
\boxed{
T_{\mathrm{total}}
\simeq
\mathcal{O}(\rho^2 n^2 d).
}
\]

\paragraph{Basic Linguistic Analysis}
In this section, we provide a preliminary linguistic analysis on real
text data.

\textbf{Example} \\
Sentence: \textit{a densely constructed, highly referential film, and an audacious return to form that can comfortably sit among jean-luc godard's finest work.} \\
Label: \textsc{Positive}

\medskip
\noindent\textbf{PLCP sparsified sentences:}

\begin{itemize}
    \item \textbf{Level 2:} \textit{referential audacious} \\
    Kept: 17.6\% (6/34) \quad Predicted: \textsc{Negative} \quad \textcolor{red}{[WRONG]}

    \item \textbf{Level 4:} \textit{referential audacious godard} \\
    Kept: 23.5\% (8/34) \quad Predicted: \textsc{Positive} \quad \textcolor{green!50!black}{[OK]}

    \item \textbf{Level 8:} \textit{referential audacious comfortably sit among jean luc godard finest} \\
    Kept: 44.1\% (15/34) \quad Predicted: \textsc{Positive} \quad \textcolor{green!50!black}{[OK]}

    \item \textbf{Level 12:} \textit{densely referential audacious return form comfortably sit among jean luc godard finest} \\
    Kept: 52.9\% (18/34) \quad Predicted: \textsc{Positive} \quad \textcolor{green!50!black}{[OK]}

    \item \textbf{Level 16:} \textit{densely highly referential film audacious return form can comfortably sit among jean luc godard s finest work} \\
    Kept: 67.6\% (23/34) \quad Predicted: \textsc{Positive} \quad \textcolor{green!50!black}{[OK]}

    \item \textbf{Level 24:} \textit{a densely constructed highly referential film and an audacious return to form that can comfortably sit among jean-luc godard's finest work.} \\
    Kept: 94.1\% (32/34) \quad Predicted: \textsc{Positive} \quad \textcolor{green!50!black}{[OK]}
\end{itemize}

In this illustrative example, sparsification first tends to remove
punctuation marks and some modifiers or function words while largely
preserving the semantic content of the sentence. However, a systematic
linguistic evaluation by a specialist would be required to draw
general conclusions and is left for future work.

\end{document}